\documentclass[journal]{IEEEtran}
\usepackage{cite}
\usepackage{url}
\usepackage{ragged2e}
\usepackage{epsfig}
\usepackage{graphicx}
\usepackage{amsmath,amssymb} 
\usepackage{subfigure}
\usepackage{algorithm}
\usepackage{algorithmicx}
\usepackage{algpseudocode}
\usepackage{graphics}
\usepackage{threeparttable}
\usepackage{color}
\usepackage[normalem]{ulem}
\usepackage{multirow}
\usepackage{float}
\usepackage{amsfonts}
\usepackage{bm}
\usepackage{array}
\usepackage[table]{xcolor}
\usepackage{colortbl}
\usepackage{pifont}
\usepackage{diagbox}
\usepackage{rotating}
\usepackage{booktabs}
\usepackage{overpic}
\usepackage{textcomp}
\usepackage{contour}
\usepackage{enumitem}
\usepackage[pagebackref=false,breaklinks=false,colorlinks=false]{hyperref}

\RequirePackage{silence}
\definecolor{mygray}{gray}{.92}

\newcommand{\tabref}[1]{Tab. \ref{#1}}

\newcommand{\figref}[1]{Fig. \ref{#1}}

\newcommand{\supp}[1]{\textcolor{magenta}{#1}}

\newcommand{\secref}[1]{$\S~$\ref{#1}}

\newcommand{\myPara}[1]{\vspace{10pt}\noindent$\bullet$~\textbf{#1} \quad}

\def\ie{\emph{i.e.}}
\def\eg{\emph{e.g.}}
\def\etc{\emph{etc}}
\def\etal{{\em et al.~}}

\def\ourmodel{\emph{BBS-Net}}

\definecolor{mygray}{gray}{.92}

\graphicspath{{./Imgs/}}
\DeclareGraphicsExtensions{.jpg,.pdf,.png}

\ifCLASSINFOpdf
\else
\fi
\begin{document}
%
\title{Bifurcated Backbone Strategy \\for RGB-D Salient Object Detection }

%
%
%
%

\author{Yingjie Zhai*, Deng-Ping Fan*, Jufeng Yang, Ali Borji, Ling Shao, \IEEEmembership{Fellow, IEEE}, \\Junwei Han, \IEEEmembership{Senior Member, IEEE}, and Liang Wang, \IEEEmembership{Fellow, IEEE}
	\IEEEcompsocitemizethanks{
		\IEEEcompsocthanksitem *Equal contribution. Listing order is random.
		Yingjie Zhai, Deng-Ping Fan and Jufeng Yang are with College of Computer Science, Nankai University. 
		Ali Borji is with Primer.AI, SF, USA. 
		Ling Shao is with the Mohamed bin Zayed University of Artificial Intelligence, Abu Dhabi, UAE, and also with the Inception Institute of Artificial Intelligence, Abu Dhabi, UAE. 
		Junwei Han is with School of Automation, Northwestern Polytechnical University, China. 
		Liang Wang is with the National Laboratory of Pattern Recognition, CAS Center for Excellence in Brain Science and Intelligence Technology, Institute of Automation, Chinese Academy of Sciences, Beijing 100190, China. 
		A preliminary version of this work has appeared in ECCV 2020~\cite{fan2020bbs}.
		Corresponding author: Jufeng Yang (yangjufeng@nankai.edu.cn).
	}
}

%
%

\markboth{IEEE TRANSACTIONS ON IMAGE PROCESSING}%
{Shell \MakeLowercase{\textit{et al.}}: Bare Demo of IEEEtran.cls for IEEE Journals}
%



\maketitle

\begin{abstract}
\justifying
Multi-level feature fusion is a fundamental topic in computer vision. It has been exploited to detect, segment and classify objects at various scales. 
When multi-level features meet multi-modal cues, the optimal feature aggregation and multi-modal learning strategy become a hot potato.
%
In this paper, we leverage the inherent multi-modal and multi-level nature of RGB-D salient object detection to devise a novel cascaded refinement network. 
In particular, first, we propose to regroup the multi-level features into teacher and student features using a bifurcated backbone strategy (BBS).
Second, we introduce a depth-enhanced module (DEM) to excavate informative depth cues from the channel and spatial views. 
Then, RGB and depth modalities are fused in a complementary way.
Our architecture, named \textbf{B}ifurcated \textbf{B}ackbone \textbf{S}trategy \textbf{Net}work (\textbf{\emph{\ourmodel}}), is simple, efficient, and backbone-independent.
Extensive experiments show that \ourmodel~significantly outperforms 18 SOTA models on 8 challenging datasets
under 5 evaluation measures, demonstrating the superiority of our approach
($\sim$4\% improvement in S-measure $vs.$ the top-ranked model: DMRA-iccv2019).
In addition, we provide a comprehensive analysis on the generalization ability of different RGB-D datasets and provide a powerful training set for future research.
\end{abstract}

\begin{IEEEkeywords}
RGB-D salient object detection, bifurcated backbone strategy, multi-level features, cascaded refinement.
\end{IEEEkeywords}

%
\IEEEpeerreviewmaketitle

\section{Introduction}

%
%
%
%
\IEEEPARstart{T}{he} goal of salient object detection (SOD) is to find and segment the most visually prominent object(s) in an image~\cite{borji2015salient,wang2019salient}.
Over the last decade, SOD has attracted significant attention due to its widespread applications in object recognition~\cite{ChengLLZRT19bing}, content-based image retrieval~\cite{cheng2017retrival}, image segmentation~\cite{Wang2015sal_seg}, image editing~\cite{Cheng2010RFA}, video analysis~\cite{Fan2019VSOD,Yan_2019_video}, and visual tracking~\cite{borji2012cvpr, Hong2015tracking}.
Traditional SOD algorithms~\cite{cheng2015GC,zhang2016co} are typically based on handcrafted features and fall short in capturing high-level semantic information (see also~\cite{borji2012state,borji2019saliency}).
Recently, convolutional neural networks (CNNs) have been used for RGB SOD~\cite{Liu2019SPBD,wang2018salient}, achieving better performance compared to the traditional methods.\par
However, the performance of RGB SOD models tends to drastically decrease in certain complex scenarios (\eg, cluttered backgrounds, multiple objects, varying illuminations, transparent objects, \etc)~\cite{chen2019TANet}.
One of the most important reasons behind these failure cases may be the lack of depth information, which is critical for saliency prediction.
For example, an object with less texture but closer to the camera is usually salient than an object with more texture but farther away.
%
Depth maps contain abundant spatial structure and layout information~\cite{piao2019DMRA}, providing geometrical cues for improving the performance of SOD.
Besides, depth information can be easily obtained using popular devices, \eg, stereo cameras, Kinect and smartphones, which are becoming increasingly more ubiquitous.
Therefore, various algorithms (\eg,~\cite{li2016saliency,zhao2019CPFP}) have been proposed to solve the SOD problem by combining RGB and depth information (\ie, RGB-D SOD).
\par
To efficiently integrate RGB and depth cues for SOD, researchers have explored different but complementary multi-modal and multi-level strategies~\cite{chen2018PCF,chen2019MMCI,li2020crossmodal} and have achieved encouraging results.
However, existing RGB-D SOD methods still have to solve the following challenges: 
\par
\begin{figure}[t!]
\centering
\begin{overpic}[width=1.0\columnwidth]{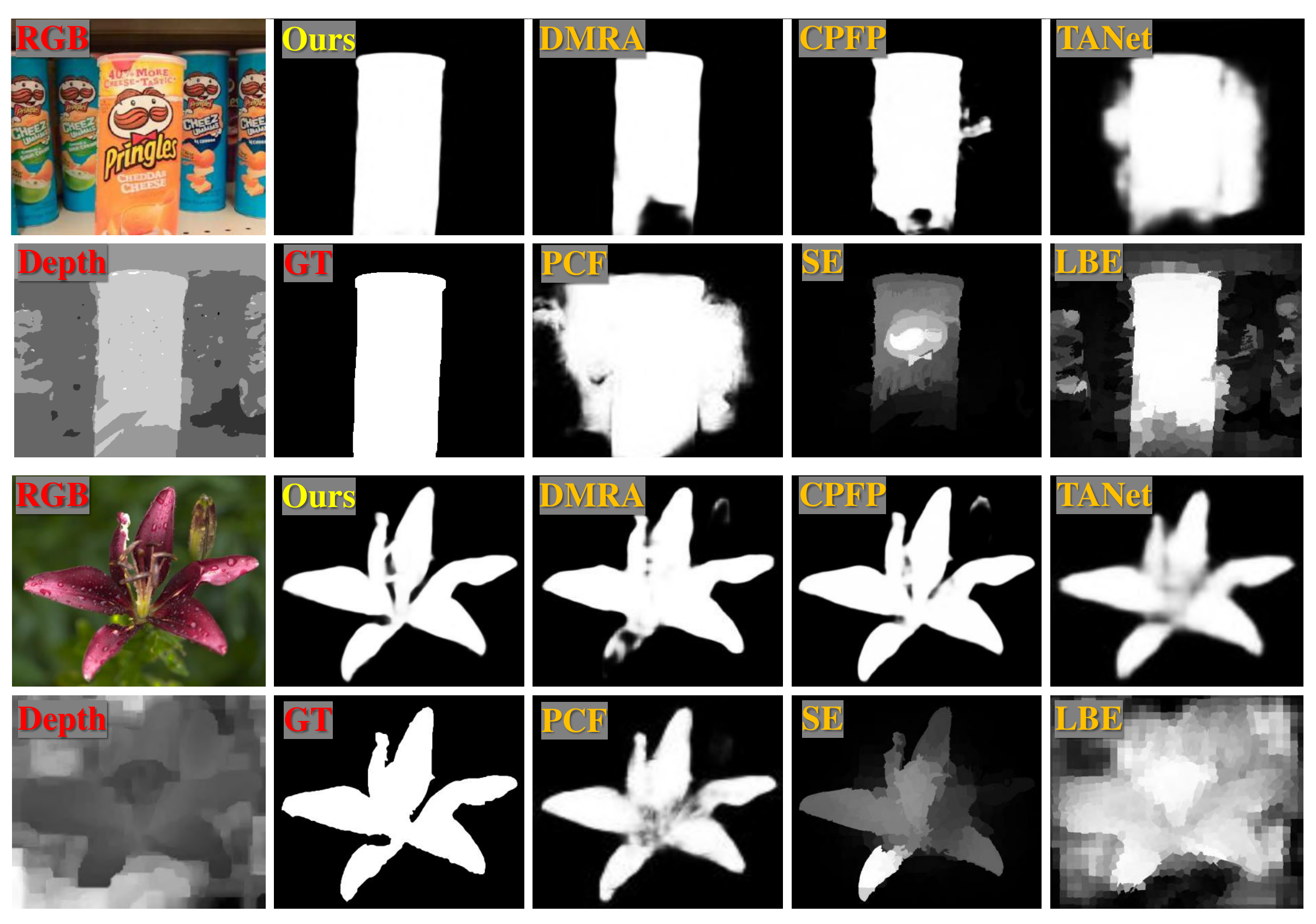}
\end{overpic}
	\caption{\small
	Saliency maps of state-of-the-art (SOTA) CNN-based methods (\ie, DMRA~\cite{piao2019DMRA}, CPFP~\cite{zhao2019CPFP}, TANet~\cite{chen2019TANet}, PCF~\cite{chen2018PCF} and Ours) and methods based on handcrafted features (\ie, SE~\cite{guo2016SE} and LBE~\cite{feng2016LBE}). Our method generates higher-quality saliency maps and suppresses background distractors in challenging scenarios (top: complex background; bottom: depth with noise).}
	\label{fig:figure1}
\end{figure}
(1) \textbf{Effectively aggregating \emph{multi-level} features.}
As discussed in \cite{Liu2019SPBD}, teacher features contain rich semantic macro information and can serve as strong guidance for locating salient objects,
while student features provide affluent micro details that are beneficial for refining object edges.
Therefore, current RGB-D SOD methods use either a dedicated aggregation strategy~\cite{piao2019DMRA,zhao2019CPFP} or a progressive merging process~\cite{LIU2019SSRC,zhu2019PDNet} to leverage multi-level features.
However, because they directly fuse multi-level features without considering level-specific characteristics, these operations suffer from the inherent problem of noisy low-level features~\cite{chen2019TANet,Wu2019CPD}.
As a result, several methods are easily confused by the background (\eg, first and second rows in \figref{fig:figure1}).\par

\begin{figure*}[t!]
	\centering
	\begin{overpic}[width=.88\linewidth]{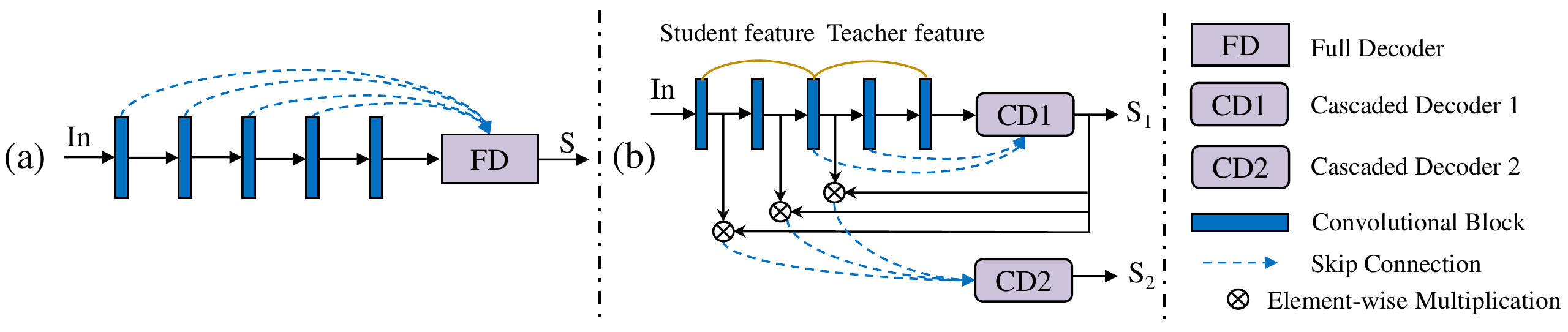}
	\end{overpic}
	\vspace{-10pt}
	\caption{
		(a) Existing multi-level feature aggregation methods for RGB-D
		SOD~\cite{chen2018PCF,zhao2019CPFP,piao2019DMRA,chen2019TANet,zhu2019PDNet,wang2019AFNet,LIU2019SSRC}.
		(b) In this paper, we adopt a bifurcated backbone strategy (BBS)
		to split the multi-level features into student and teacher features.
		The initial saliency map $S_1$ is utilized to refine the student features to effectively suppress distractors. Then, the refined features are 
		passed to another cascaded decoder to generate the final saliency map $S_2$.
	}
	\label{fig:FrameworkCompare}
\end{figure*}

(2) \textbf{Excavating informative cues from the \emph{depth modality}.}
Previous algorithms usually regard the depth map as a fourth-channel input~\cite{cong2019HSCS,peng2014LHM} of the original three-channel RGB image, or fuse RGB and depth features by simple summation~\cite{fan2014DSP,fang2014TIP} and multiplication~\cite{cheng2014DESM,zhu2017CDCP}.
However, these methods treat depth and RGB information from the same perspective and ignore the fact that RGB images capture color and texture, whereas depth maps capture the spatial relations among objects.
Due to this modality difference, the above-mentioned simple combination methods are not very efficient.
Further, depth maps often have low quality, which introduces randomly distributed errors and redundancy into the network~\cite{fan2019D3Net}.
For example, the depth map in the last row of \figref{fig:figure1} is blurry and noisy. As a result, many methods (\eg, the top-ranked model DMRA\cite{piao2019DMRA}) fail to detect the full extent of the salient object.
\par

To address the above issues, we propose a novel \textbf{B}ifurcated \textbf{B}ackbone \textbf{S}trategy \textbf{Net}work (\ourmodel) for RGB-D SOD.
The proposed method exploits multi-level features in a cascaded refinement way to suppress distractors in the lower layers.
This strategy is based on the observation that teacher features provide discriminative semantic information without redundant details~\cite{Liu2019SPBD,Wu2019CPD}, which may contribute significantly to eliminating the lower-layer distractors.
As shown in \figref{fig:FrameworkCompare} (b), \ourmodel~contains two cascaded decoder stages:
(1) Cross-modal teacher features are integrated by the first cascaded decoder $CD1$ to predict an initial saliency map $S_{1}$.
(2) Student features are refined by an
element-wise multiplication with the initial saliency map $S_{1}$ and are then aggregated
by another cascaded decoder $CD2$ to produce the final saliency map $S_{2}$.
To fully capture the informative cues in the depth map and improve the compatibility of RGB and depth features, we further introduce a depth-enhanced module (DEM). This module exploits the inter-channel and spatial relations of the depth features and discovers informative depth cues.
%
%

%
Additionally, to obtain reasonable performance in real-world scenarios, not only an efficient model is needed but also a dataset with great generalization ability is required to train such model.
There are several large-scale RGB-D datasets, \eg, NJU2K~\cite{ju2014ACSD}, NLPR~\cite{peng2014LHM}, STERE~\cite{niu2012STERE},
SIP~\cite{fan2019D3Net} and DUT~\cite{piao2019DMRA} with more than $1,000$ image pairs.
However, researchers have often trained RGB-D models on the fixed training set (\ie, $1,485$ images from NJU2K and $700$ images from NLPR). This limits the model's generation ability in various scenarios.
Further, they have not studied the generalization ability of different datasets and have not proposed powerful training sets.
%
In this paper, one of our goals is to study this problem in detail.

Our main contributions are summarized as follows:

\begin{itemize}
	\item
	\textbf{We propose a powerful Bifurcated Backbone Strategy Network (\ourmodel)} to deal with multiple complicated real-world scenarios in RGB-D SOD.
	To address the long-overlooked problem of noise in low-level features decreasing the performance of saliency models, 
	we carefully explore the characteristics of multi-level features in a bifurcated backbone strategy (BBS), \ie, features are split into two groups, as shown in \figref{fig:FrameworkCompare} (b).
	In this way, noise in student features can be eliminated effectively by the saliency map generated from teacher features.
	
	\item	
	\textbf{We further introduce a depth-enhanced module (DEM)} in \ourmodel~to enhance the depth features before merging them with the RGB features.
	The DEM module concentrates on the most informative parts of depth maps by two sequential attention operations. 
	We leverage the attention mechanism to excavate important cues from the depth features of multiple side-out layers. 
	This module is simple but has proven effective for fusing RGB and depth modalities in a complementary way.
	
	\item
	\textbf{We conduct a comprehensive comparison with $18$ SOTA methods} using various metrics (\eg, max F-measure, MAE, S-measure, max E-measure, and PR curves).
	Experimental results show that \ourmodel~outperforms all of these methods on eight public datasets, by a large margin.
	In terms of the predicted saliency maps, \ourmodel~generates maps with sharper edges and fewer background distractors compared to existing models.
	%
	
	\item
	\textbf{We conduct a number of cross-dataset experiments} to evaluate the quality of current popular RGB-D datasets and introduce a training set with high generalization ability for fair comparison and future research.
	Current RGB-D methods train their networks using the fixed training-test splits of different datasets, without exploring the difficulties of those datasets.
	To the best of our knowledge, we are the first to investigate this important but overlooked problem in the area of RGB-D SOD.
\end{itemize}

This work is based on our previous conference paper~\cite{fan2020bbs} and extends it significantly in five ways:
1) We further extend the approach by designing a depth adapter module, which makes the model contain around $50$ percent parameters of the previous version but with similar performance.
2) We provide more details and experiments regarding our \ourmodel~model, including motivation, feature visualizations, experimental settings, \etc.
3) We investigate several previously unexplored issues, including cross-dataset generalization ability, post-processing methods, failure cases analysis, \etc.
4) To further demonstrate our model performance, we conduct several comprehensive experiments over the recently released dataset, DUT~\cite{piao2019DMRA}. 
5) We perform in-depth analyses and draw several novel conclusions which are critical in developing more powerful models in the future. 
We are hopeful that our study will provide deep insights into the underlying design mechanisms of RGB-D SOD, and will spark novel ideas.
The complete algorithm, benchmark results, and post-processing toolbox are publicly available at \supp{\href{https://github.com/zyjwuyan/BBS-Net}{https://github.com/zyjwuyan/BBS-Net}.}

\vspace{-5pt}
\section{Related Works}\label{sec:related_works}

\vspace{-5pt}
\subsection{Salient Object Detection}\label{sec:sod}
\vspace{-5pt}
Over the past several decades, SOD~\cite{liu2010learning,achanta2009frequency,fan2018foreground} has garnered significant research interest due to its diverse applications~\cite{LiY16,ZhangWLWY17,zhao2020suppress}.
In early years, SOD methods were primarily based on intrinsic prior knowledge such as center-surround color contrast~\cite{itti1998model}, global region contrast~\cite{cheng2015GC}, background prior~\cite{li2015visual} and appearance similarity~\cite{cheng2013efficient}.
However, these methods heavily rely on heuristic saliency cues and low-level handcrafted features, thus lacking the guidance of high-level semantic information.\par 

Recently, to solve this problem, deep learning based methods~\cite{Chen_2018_reverse,Zhang2018PAGR,Su_2019_ICCV,zhang2020multi,Li_2019_video} have been explored, exceeding handcrafted feature-based methods in complex scenarios.
These deep methods~\cite{ZhangWLWR17} usually leverage CNNs to extract multi-level multi-scale features from RGB images and then aggregate them to predict the final saliency map.
Such multi-level multi-scale features~\cite{wang2019progressive,wei2019f3net} can help the model better understand the contextual and semantic information to generate high-quality saliency maps.
Besides, since image-based SOD may be limited in some real-world applications such as video captioning~\cite{PanYLM17caption}, autonomous driving~\cite{ZhangFU16drive} and robotic interaction~\cite{XuPCYH16interactive}, SOD algorithms~\cite{Fan2019VSOD,Yan_2019_video} have also been explored for video analysis. 

To further overcome the limits of deep models, researchers have also proposed to excavate edge information~\cite{XieT17edge} to guide prediction.
These methods use an auxiliary boundary loss to improve the training and representative ability of segmentation tasks~\cite{ZhugeYZL18edge,zhao2019egnet,wu2019stacked}.
With the auxiliary guidance from the edge information, deep models can predict maps with finer and sharper edges.
In addition to edge guidance, another useful type of auxiliary information are depth maps, which capture the spatial distance information. These are the main focus of this paper. 

\subsection{RGB-D Salient Object Detection}\label{sec:rgbd}
%
%

\myPara{\textbf{Traditional Models.}} Previous algorithms for RGB-D SOD mainly rely on extracting handcrafted features~\cite{cheng2014DESM, zhu2017CDCP} from RGB and depth images.
Contrast-based cues, including edge, color, texture and region, are largely utilized by these methods to compute the saliency of a local region.
For example, Desingh~\etal\cite{desingh2013BMCV} adopted the region-based contrast to calculate contrast strengths for the segmented regions.
Ciptadi~\etal\cite{ciptadi2013BMCV} used surface normals and color contrast to compute saliency. 
However, the local contrast methods are easily disturbed by high-frequency content~\cite{qu2017DF}, since they mainly rely on the boundaries of salient objects.
Therefore, some algorithms, such as spatial prior~\cite{cheng2014DESM}, global contrast~\cite{cong2019DGTM}, and background prior~\cite{Shigematsu2017BED}, proposed to compute saliency by combining both local and global information.\par
\par
To combine saliency cues from RGB and depth modalities more effectively, researchers have explored multiple fusion strategies.
Some methods~\cite{peng2014LHM,cong2019HSCS} process RGB and depth images together by regarding depth maps as fourth-channel inputs (early fusion).
This operation is simple but does not achieve reliable results, since it disregards the differences between the RGB and depth modalities.
Therefore, some algorithms~\cite{fan2014DSP,zhu2017CDCP} extract the saliency information from the two modalities separately by first leveraging two backbones to predict saliency maps and then fusing the saliency results (late fusion).
Besides, to enable the RGB and depth modalities to share benefits, other methods~\cite{feng2016LBE,ju2014ACSD} fuse RGB and depth features in a middle stage and then produce the corresponding saliency maps (middle fusion).
%
Deep models also use the above three fusion strategies, and our method falls under the middle fusion category.

\begin{figure*}[t!]
	\small
	\centering
	\begin{overpic}[width=0.98\linewidth]{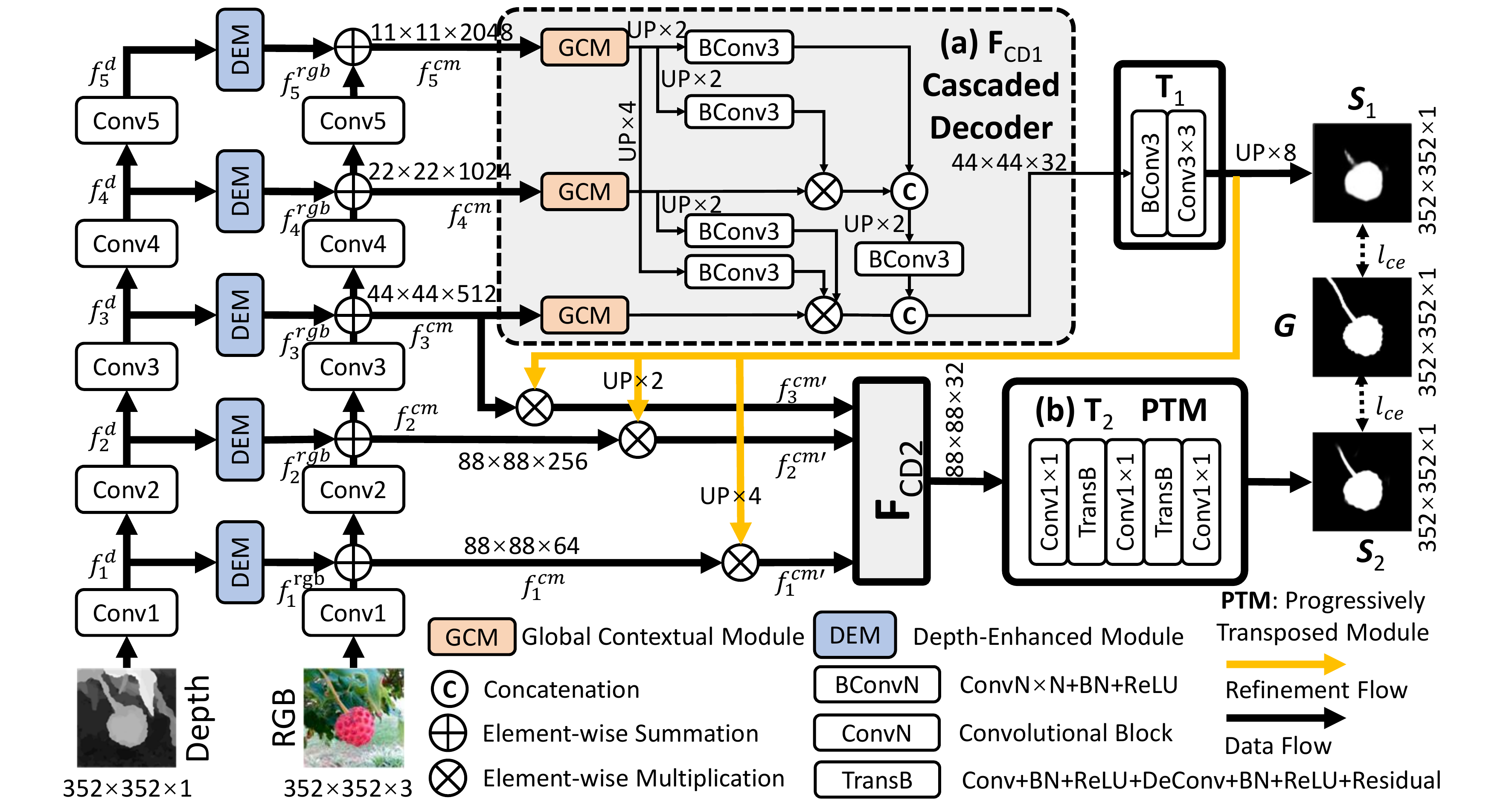}
	\end{overpic}
	\vspace{-5pt}
	\caption{ \textbf{Architecture of our \ourmodel.}
		Feature Extraction: `\textit{Conv1}'$\sim$`\textit{Conv5}' denote different layers from ResNet-50~\cite{He2016resnet}.
		Multi-level features ($f_1^d\sim f_5^d$) from the depth branch are enhanced by the DEM and are
		then fused with features (\ie, $f_1^{rgb}\sim f_5^{rgb}$) from the RGB branch.
		Stage 1: cross-modal teacher features ($f_3^{cm}\sim f_5^{cm}$) are first aggregated by the cascaded decoder (a)
		to produce the initial saliency map $S_1$.
		Stage 2: Then, student features ($f_1^{cm}\sim f_3^{cm}$) are refined by the initial saliency map $S_1$
		and are integrated by another cascaded decoder to predict the final saliency map $S_2$.
		See \secref{sec:proposedMethod} for details. }
	\label{fig:pipe}
\end{figure*}

\myPara{\textbf{Deep Models.}}
Early deep methods~\cite{qu2017DF,Shigematsu2017BED} compute saliency confidence scores by first extracting handcrafted features, and then feeding them to CNNs.
However, these algorithms need the low-level handcrafted features to be manually designed as input, and thus cannot be trained in an end-to-end manner.
More recently, researchers have begun to extract deep RGB and depth features using CNNs in a bottom-up fashion~\cite{han2018CTMF}.
Unlike handcrafted features, deep features contain a lot of contextual and semantic information, and can thus better capture representations of the RGB and depth modalities.
These methods have achieved encouraging results, which can be attributed to two important aspects of feature fusion.
One is their extraction and fusion of multi-level and multi-scale features from different layers, while the other is the mechanism by which the two different modalities (RGB and depth) are combined.\par

Various architectures have been designed to effectively integrate the multi-scale features.
For example, Liu \etal\cite{LIU2019SSRC} obtained saliency map outputs from each side-out features by feeding a four-channel RGB-D image into a single backbone (single stream).
Chen \etal\cite{chen2018PCF} leveraged two independent networks to extract RGB and depth features respectively, and then combined them in a progressive merging way (double stream).
Furthermore, to learn supplementary features,~\cite{chen2019TANet} designed a three-stream network consisting of two modality-specific streams and a parallel cross-modal distillation stream to exploit complementary cross-modal information in the bottom-up feature extraction process (three streams).
Depth maps are sometimes low-quality and may thus contain significant noise or misleading information, which greatly decreases the performance of SOD models.
To address this issue, Zhao~\etal\cite{zhao2019CPFP} proposed a contrast-enhanced network to improve the quality of depth maps using the contrast prior.
Fan \etal\cite{fan2019D3Net} designed a depth depurator unit to evaluate the quality of depth maps and filter out the low-quality ones automatically.
Three recent works have explored uncertainty~\cite{Zhang2020UCNet}, depth prediction~\cite{wang2020synergistic} and a joint learning strategy~\cite{Fu2020JLDCF} for saliency detection and achieved reasonable performance.
There were also some concurrent works published in recent top conferences (\eg, ECCV~\cite{Luo_2020_ECCV,zhao2020single,li2020rgbd}).
Discussing these works in detail is beyond the scope of this article. Please refer to the online benchmark (\supp{\href{http://dpfan.net/d3netbenchmark/}{http://dpfan.net/d3netbenchmark/}}) and the latest survey~\cite{zhou2020rgbd} for more details.

%
%

\section{Proposed Method}\label{sec:proposedMethod}
\subsection{Overview}\label{sec:overview}
Current popular RGB-D SOD models directly integrate multi-level features using a single decoder (\figref{fig:FrameworkCompare} (a)).
In contrast, the network flow of the proposed \ourmodel~(\figref{fig:pipe}) explores a bifurcated backbone strategy.
In \secref{sec:BifucatedStrategy}, we first detail the proposed bifurcated backbone strategy with the cascaded refinement mechanism. Then, to fully excavate informative cues from the depth map, we introduce a new depth-enhanced module in \secref{sec:DEM}.
Additionally, we design a depth adapter module to further improve the efficiency of the model in \secref{sec:efficiency}.

\subsection{Bifurcated Backbone Strategy (BBS)}\label{sec:BifucatedStrategy}

%
Our cascaded refinement mechanism leverages the rich semantic information in high-level cross-modal features to suppress background distractors.
To support such a feat, we devise a bifurcated backbone strategy (BBS). It divides the multi-level cross-modal features into two groups, \ie, $\textbf{G}_1$ = \{\textit{Conv1, Conv2, Conv3}\} and $\textbf{G}_2 = $\{\textit{Conv3, Conv4, Conv5}\}, where \textit{Conv3} is the split point.
The original multi-scale information is well preserved by each group.

\myPara{\textbf{Cascaded Refinement Mechanism.}}
To effectively leverage the characteristics of the features in the two groups' features, we train the network using a cascaded refinement mechanism.
This mechanism first generates an initial saliency map with three cross-modal teacher features (\ie, $\textbf{G}_2$) and then enhances the details of the initial saliency map $S_1$ with three cross-modal student features (\ie, $\textbf{G}_1$), which are refined by the
initial saliency map. 
%
%
This is based on the observation that high-level features contain rich semantic information that helps locate salient objects, while low-level features provide micro-level details that are beneficial for refining the boundaries.
In other words, by exploring the characteristics of the multi-level features, this strategy can efficiently suppress noise in low-level cross-modal features, and can produce the final saliency map through a progressive refinement. \par
%
Specifically, we first merge RGB and depth features processed by the DEM to obtain the cross-modal features $\{f_i^{cm}; i=1,2,...,5\}$.
In stage one, the three cross-modality teacher features (\ie, $f_3^{cm},f_4^{cm},f_5^{cm}$) are aggregated by the first cascaded decoder, which is denoted as:
\begin{equation}\label{equ:S1}
S_1=\textbf{T}_1\big(\textbf{F}_{CD1}(f_3^{cm},f_4^{cm},f_5^{cm})\big),
\end{equation}
where $\textbf{F}_{CD1}$ is the first cascaded decoder, $S_1$ is the initial saliency map, and $\textbf{T}_1$ represents two simple convolutional layers that transform the channel number from $32$ to $1$.
In stage two, we leverage the initial saliency map $S_1$ to refine the three cross-modal student features, which is defined as:
\begin{equation}\label{equ:refine}
f_i^{cm^{\prime}}=f_i^{cm}+f_i^{cm}\odot S_1,
\end{equation}
where $f_i^{cm^{\prime}}$ ($i\in\{1,2,3\}$) represents the refined features and $\odot$ denotes the element-wise multiplication.
After that, the three refined student features are aggregated by another decoder followed by a progressively transposed module (PTM), which is formulated as:
\begin{equation}\label{equ:S2}
S_2=\textbf{T}_2\Big(\textbf{F}_{CD2}(f_1^{cm^{\prime}},f_2^{cm^{\prime}},f_3^{cm^{\prime}})\Big),
\end{equation}
where $\textbf{F}_{CD2}$ is the second cascaded decoder, $S_2$ denotes the final saliency map, and $\textbf{T}_2$ represents the PTM module. 
%
%

\myPara{\textbf{Cascaded Decoder.}}
After computing the two groups of multi-level cross-modal features ($\{f_i^{cm},f_{i+1}^{cm},f_{i+2}^{cm}\},i\in\{1,3\}$), which are a fusion of the RGB and depth features from multiple layers, we need to efficiently leverage the multi-scale multi-level information in each group to carry out the cascaded refinement.
Therefore, we introduce a light-weight cascaded decoder~\cite{Wu2019CPD} to integrate the two groups of multi-level cross-modal features.
As shown in \figref{fig:pipe} (a), the cascaded decoder consists of three global context modules (GCM) and a simple feature aggregation strategy.
The GCM is refined from the RFB module~\cite{liu2018rfb}.
Specifically, it contains an additional branch to enlarge the receptive field and a residual connection~\cite{He2016resnet} to preserve the information.
The GCM module thus includes four parallel branches.
For all of these branches, a $1\times 1$ convolution is first applied to reduce the channel size to $32$.
Then, for the $k^{th}$ ($k\in \{2,3,4\}$) branch, a convolution operation with a kernel size of $2k-1$ and dilation rate of 1 is applied.
This is followed by another $3\times3$ convolution operation with the dilation rate of $2k-1$.
We aim to excavate the global contextual information from the cross-modal features.
Next, the outputs of the four branches are concatenated together and a 3$\times$3 convolution operation is then applied to reduce the channel number to 32.
Finally, the concatenated features form a residual connection with the input features.
The GCM module operation in the two cascaded decoders is denoted by:
\begin{equation}\label{equ:gcm}
f_i^{gcm}=\textbf{F}_{GCM}(f_i).
\end{equation} 
To further improve the representations of cross-modal features,
we leverage a pyramid multiplication and concatenation feature aggregation strategy
to aggregate the cross-modal features ($\{f_i^{gcm},f_{i+1}^{gcm},f_{i+2}^{gcm}\},i\in\{1,3\}$).
As illustrated in \figref{fig:pipe} (a), first, each refined feature $f_i^{gcm}$ is updated by multiplying it with all higher-level features:
\begin{equation}\label{equ:pyramid_multipy}
f_i^{gcm^{\prime}}=f_i^{gcm} \odot \Pi_{k=i+1}^{k_{max}} Conv\Big(\textbf{F}_{U P}\left(f_k^{gcm}\right)\Big),
\end{equation}
in which $i \in \{1,2,3\}$, $k_{max}=3$ or $ i \in \{3,4,5\}$, $k_{max}=5$. $\textbf{F}_{UP}$ represents the upsampling operation if the features are not of the same scale. $\odot$ represents the element-wise multiplication, and $Conv (\cdot)$ represents the standard 3$\times$3 convolution operation. Then, the updated features are integrated by a progressive concatenation strategy to produce the output:
\begin{equation}\label{equ:agg}
S=\textbf{T}\left(\left[f_k^{gcm^{\prime}};Conv\Big(\textbf{F}_{UP}\left[f_{k+1}^{gcm^{\prime}};Conv\left(\textbf{F}_{UP}(f_{k+2}^{gcm^{\prime}})\Big)\right]\right)\right]\right),
\end{equation}
where $S$ is the predicted saliency map, $[x;y]$ denotes the concatenation operation of $x$ and $y$, and $k\in\{1,3\}$. In the first stage, $\textbf{T}$ denotes two sequential convolutional layers (\ie, $\textbf{T}_1$), while, for the second stage, it represents the PTM module (\ie, $\textbf{T}_2$). 
The scale of the output of the second decoder is 88$\times $88, which is $1/4$ of the ground-truth (352$\times$352), so directly upsampling the output to the size of the ground-truth will lose some details.
To address this issue, we propose a simple yet effective progressively transposed module (PTM, \figref{fig:pipe} (b)) to generate the final predicted map ($S_2$) in a progressive upsampling way.
It consists of two residual-based transposed blocks~\cite{2019huAcnet} and three sequential $1\times 1$ convolutions.
Each residual-based transposed block contains a $3\times3$ convolution and a residual-based transposed convolution.

Note that the proposed cascaded refinement mechanism is different from the recent refinement strategies CRN~\cite{chen2017photographic}, SRM~\cite{WangBZZL17}, R3Net~\cite{deng2018r3net}, and RFCN~\cite{wang2018salient} in its usage of the initial map and multi-level features.
The obvious difference and advantage of the proposed design is that our model only requires one round of saliency refinement to produce a good saliency map, while CRN, SRM, R3Net, and RFCN all need more iterations, which increases both the training time and computational resources.
Besides, the proposed cascaded mechanism is also different from CPD~\cite{Wu2019CPD} in that it exploits both the details in student features and the semantic information in teacher features, while suppressing the noise in the student features at the same time. 
\subsection{Depth-Enhanced Module (DEM)}\label{sec:DEM}
To effectively fuse the RGB and depth features, two main problems need to be solved: a) the compatibility of RGB and depth features needs to be improved due to the intrinsic modality difference, and b) the redundancy and noise in low-quality depth maps must be reduced.
Inspired by~\cite{Woo2018CBAM}, we design a depth-enhanced module (DEM) to address the issues by improving the compatibility of multi-modal features and excavating informative cues from the depth features. 

\par
Specifically, let $f_i^{rgb}$, $f_i^{d}$ represent the feature maps of the $i^{th}$ ($i\in 1,2,...,5$) side-out layer from the RGB and depth branches, respectively.
As shown in \figref{fig:pipe}, each DEM is added before each side-out feature map from the depth branch to enhance the compatibility of the depth features.
This side-out process improves the saliency representation of depth features and, at the same time, preserves the multi-level multi-scale information.
The fusion process of the two modalities is depicted as:
\begin{equation}\label{equ:dem_fuse}
f_i^{cm}=f_i^{rgb}+\textbf{F}_{DEM}(f_i^{d}),
\end{equation}
where $f_i^{cm}$ denotes the cross-modal features of the $i^{th}$ layer.
The DEM module contains a sequential channel attention operation and a spatial attention operation, which are formulated as:
\begin{equation}\label{equ:dem_cs_att}
\textbf{F}_{DEM}(f_i^{d})=\textbf{S}_{att}\Big(\textbf{C}_{att}(f^{d}_i)\Big),
\end{equation}
in which $\textbf{C}_{att}(\cdot)$ and $\textbf{S}_{att}(\cdot)$ represent the spatial and channel attention operations, respectively. More specifically, the channel attention is implemented as:
\begin{equation}\label{equ:dem_c_att}
\textbf{C}_{att}(f)=\textbf{M}\Big(\textbf{P}_{max}(f)\Big)\otimes f,
\end{equation}
where $\textbf{P}_{max}(\cdot)$ denotes the global max pooling operation for each feature map, $\textbf{M} (\cdot)$ represents a multi-layer (two-layer) perceptron, $f$ denotes the input feature map, and $\otimes$ is the multiplication by the dimension broadcast.
The spatial attention is denoted as:
\begin{equation}\label{equ:S_att}
\textbf{S}_{att}(f)=Conv\Big(\mathbf{R}_{max}(f)\Big)\odot f,
\end{equation}
where $\mathbf{R}_{max}(\cdot)$ is the global max pooling operation for each point in the feature map along the channel axis.
The proposed depth enhanced module is different from previous RGB-D algorithms, which fuse the multi-level cross-modal features
by direct concatenation~\cite{chen2018PCF,chen2019TANet,zhu2019PDNet}, enhance the multi-level depth features by a simple convolutional layer~\cite{piao2019DMRA} or improve the depth map by contrast prior~\cite{zhao2019CPFP}.
To the best of our knowledge, we are the first to introduce the attention mechanism to excavate informative cues from depth features in multiple side-out layers.
Our experiments (see \tabref{tab:ablation} and \figref{fig:visual_ablation}) demonstrate the effectiveness of our approach in
improving the compatibility of multi-modal features.

Besides, the spatial and channel attention mechanisms are different from the operation proposed in~\cite{Woo2018CBAM}.
Based on the fact that SOD aims at finding the
most prominent objects in an image, we only leverage a single global max pooling~\cite{oquab2015GMP} to excavate the most critical cues in depth
features, which reduces the complexity of the module.
\begin{figure}[t!]
	\small
	\centering
	\begin{overpic}[width=0.8\linewidth]{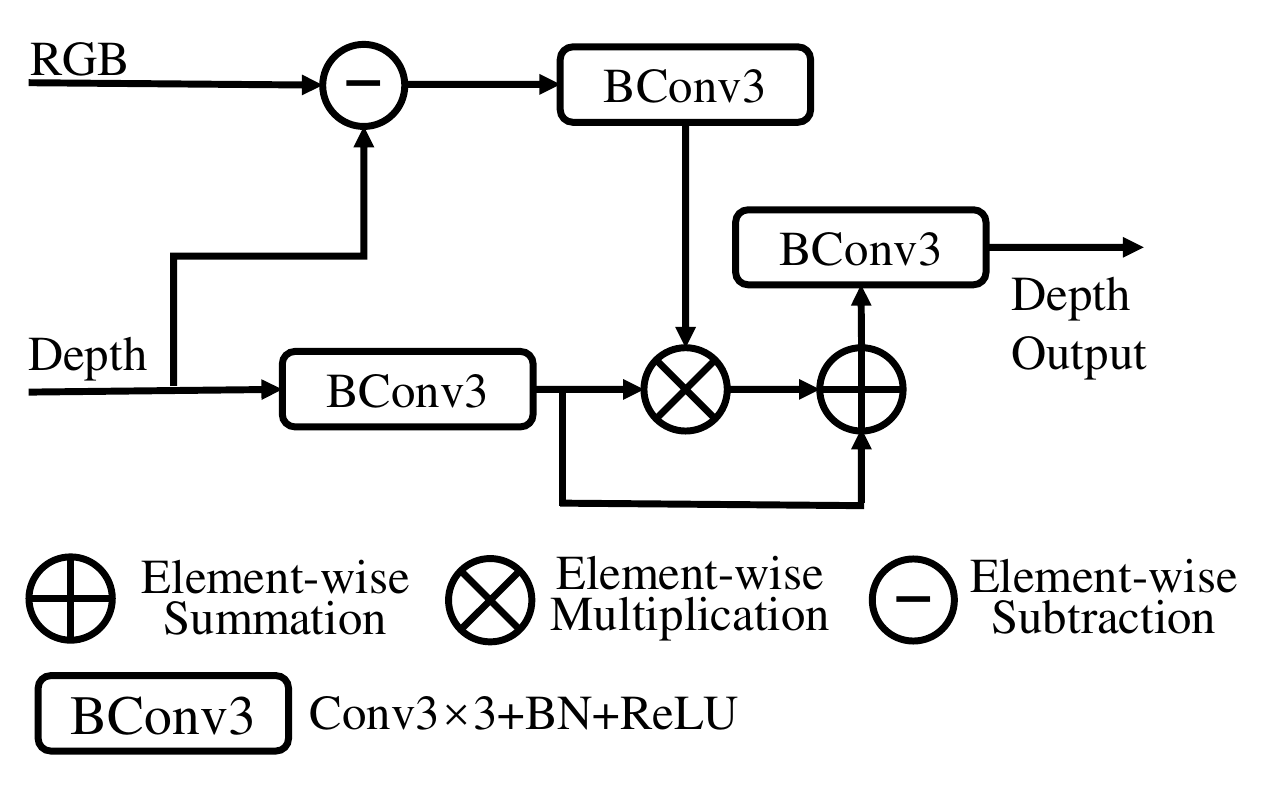}
	\end{overpic}
	\vspace{-5pt}
	\caption{ \textbf{Architecture of the depth adapter module (DAM).}
	 }
	\label{fig:dam}
\end{figure}
\subsection{Improve the efficiency of BBS-Net.}\label{sec:efficiency}
Note that the above proposed~\ourmodel~leverages two backbones, without sharing weights, to extract RGB features and depth features.
Such a design can make the model extract discriminative RGB features and depth features, respectively, but also introduces more parameters, leading to a suboptimal solution for lightweight applications.
However, making the two branches share weights can cause a big degradation of the performance.
It may be because the RGB image and the depth image are two different modalities, \ie, RGB image contains color, structure, and semantic information while the depth image includes the spatial distance information.
Thus a naive sharing-weight mechanism of the two-branch backbones cannot be suitable to extract the multi-modal features.
To solve this problem, we design a depth adapter module (DAM) to consider the modality difference of the RGB image and depth image. The same backbone can be suitable to extract two-modality features without decreasing much performance.
\par
The whole architecture of the DAM is shown in \figref{fig:dam}.
Let $I_{rgb}$ and $I_{depth}$ denote the input RGB and depth image pair, respectively.
We first calculate the modality difference $I_{dif}$ by,
\begin{equation}\label{equ:S1}
I_{dif}=Conv(I_{rgb}-I_{depth}),
\end{equation}
where $I_{depth}$ is broadcast to the same dimension as $I_{rgb}$. 
Such an operation can make the model understand the explicit difference between the depth image and the RGB image.
Then the adapted depth output is computed by:
\begin{equation}\label{equ:S1}
I^{\prime}_{dif}=Conv\Big(Conv(I_{depth})+Conv(I_{depth})*I_{dif}\Big).
\end{equation}
In the efficient version of BBS-Net, the backbones of the two branches share parameters.
When calculating the depth features, the depth image is first fed to the DAM module to obtain the adapted depth information and is then fed to the backbone to extract features.
To further reduce model parameters, we also remove the last progressively transposed module (which makes negligible performance degradation) in the efficient version of \ourmodel.
\begin{figure*}[t!]
	\small
	\centering
	\begin{overpic}[width=.98\linewidth]{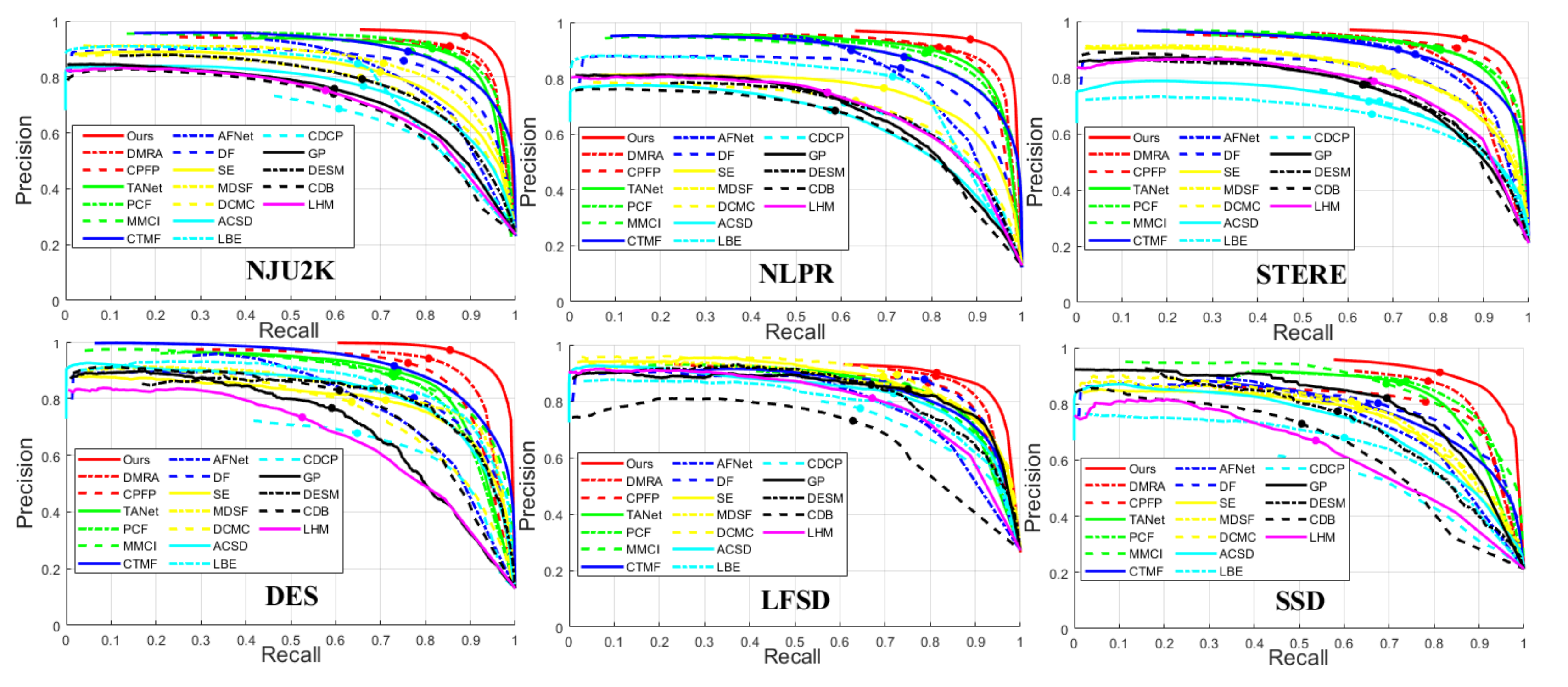}
		
	\end{overpic}
	\vspace{-10pt}
	\caption{ \textbf{PR Curves of the proposed model and 18 SOTA algorithms over six datasets.} Dots on the curves represent the value of precision and recall at the maximum F-measure.}
	\label{fig:prcurve}
\end{figure*}

\subsection{Implementation Details}\label{sec:implement}
\myPara{\textbf{Training Loss.}} Let $\mathcal{H}$ and $\mathcal{W}$ denote the height and width of the input images. Given the input RGB image $X\in R^{\mathcal{H}\times\mathcal{W}\times3}$ and its corresponding depth map $D\in R^{\mathcal{H}\times\mathcal{W}\times1}$, our model predicts an initial saliency map $S_1\in [0,1]^{\mathcal{H}\times\mathcal{W}\times1}$ and a final saliency map $S_2\in [0,1]^{\mathcal{H}\times\mathcal{W}\times1}$.
Let $G\in \{0,1\}^{\mathcal{H}\times\mathcal{W}\times1}$ denote the binary ground-truth saliency map. We jointly optimize the two cascaded stages by defining the total loss:
\begin{equation}\label{equ:totalloss}
\mathcal L=\alpha \ell_{ce}(S_1,G)+ (1-\alpha) \ell_{ce}(S_2,G),
\end{equation}
in which $\ell_{ce}$ represents the binary cross entropy loss~\cite{zhao2019CPFP} and $\alpha\in [0,1]$ controls the trade-off between the two parts of the losses. The $\ell_{ce}$ is computed as:
\begin{equation}\label{equ:celoss}
\ell_{ce}(S,G)=G\log S + (1-G)\log(1-S),
\end{equation}
where $S$ is the predicted saliency map.\par
\myPara{\textbf{Training and Test Protocol.}}
We use PyTorch~\cite{steiner2019pytorch} to implement our model on a single 1080Ti GPU. 
Parameters of the backbone network (ResNet-50~\cite{He2016resnet}) are initialized from the model pre-trained on ImageNet~\cite{KrizhevskySH2012Imagenet}.
Other parameters are initialized using the default PyTorch settings.
We discard the last pooling and fully connected layers of ResNet-50 and
leverage each middle output of the five convolutional blocks as the side-out feature maps.
The two branches do not share weights and the only difference between them is that the
depth branch has the input channel number set to one.
Note for the efficient version \ourmodel$^{\divideontimes}$, the two branches share weights, and the depth images are first fed to the depth adapter module to reduce the modality difference.
The Adam algorithm~\cite{KingmaB2014adam} is used to optimize our model, the betas are set to $0.9$ and $0.99$, and the weight decay is set to $0$.
We set the initial learning rate to 1e-4 and divide it by $10$ every 60 epochs.
The gradients are clipped into $[-0.5,0.5]$ to make the training stable.
The input RGB and depth images are resized to $352\times352$ for both the training and test phases.
We augment all the training images using multiple strategies (\ie, random flipping, rotating, and border clipping).
It takes about ten hours to train the model with a mini-batch size of 10 for $150$ epochs.
Our experiments show that the model is robust to the hyper-parameter $\alpha$.
Thus, we set $\alpha$ to $0.5$ (\ie, same importance for the two losses).
In the test phase, the predicted maps are upsampled to the same dimension of ground truth by the bilinear interpolation and are then normalized to [0,1].
%
%
\section{Experiments and Results}
%
%

\subsection{Experimental Settings}\label{sec:experimental_settings}

\myPara{\textbf{Datasets.}}\label{sec:dataset}
We conduct our experiments on eight challenging RGB-D SOD benchmark datasets:
NJU2K~\cite{ju2014ACSD}, NLPR~\cite{peng2014LHM}, STERE~\cite{niu2012STERE}, DES~\cite{cheng2014DESM}, LFSD~\cite{li2014LFSD}, SSD~\cite{zhu2017SSD},
SIP~\cite{fan2019D3Net} and DUT~\cite{piao2019DMRA}.
%
%
\textbf{NJU2K}~\cite{ju2014ACSD} is the largest RGB-D dataset containing $1,985$ image pairs. 
\textbf{NLPR}~\cite{peng2014LHM} consists of $1,000$ image pairs captured by a standard \textit{Microsoft Kinect} with a resolution of $640\times480$. 
\textbf{STERE}~\cite{niu2012STERE} is the first stereoscopic photo collection, containing $1,000$ images downloaded from the Internet. 
\textbf{DES}~\cite{cheng2014DESM} is a small-scale RGB-D dataset that includes $135$ indoor image pairs. 
\textbf{LFSD}~\cite{li2014LFSD} contains $60$ image pairs from indoor scenes and $40$ image pairs from outdoor scenes. 
\textbf{SSD}~\cite{zhu2017SSD} includes $80$ images picked from three stereo
movies with both indoor and outdoor scenes. The collected images have a high resolution of $960\times 1,080$.
\textbf{SIP}~\cite{fan2019D3Net} consists of $1,000$ image pairs captured by a \textit{smart phone} with a resolution of $992\times 744$, using a dual camera. 
\textbf{DUT}~\cite{piao2019DMRA} 
includes $1200$ images from multiple challenging scenes (\eg, transparent objects, multiple objects, complex backgrounds and low-intensity environments). 
\begin{table}[t!]
	\caption{ Performance of different models on the DUT~\cite{piao2019DMRA} dataset. Models are trained and tested on the DUT using the proposed training and test sets split from~\cite{piao2019DMRA}.}
	\vspace{-5pt}
	\label{tab:dut_result}
	\centering
	\setlength\aboverulesep{0.5pt}\setlength\belowrulesep{1pt}
	\renewcommand{\arraystretch}{0.5}
	\setlength{\tabcolsep}{2.0mm}
	\footnotesize
	\begin{tabular}{c|r|ccc}
		\toprule
		\multicolumn{1}{c|}{\multirow{2}{*}{\#}}&\multicolumn{1}{c|}{\multirow{2}{*}{\diagbox{Method}{Dataset}}} & \multicolumn{3}{c}{DUT~\cite{piao2019DMRA}}\\\cline{3-5}
		&&$S_{\alpha}\uparrow$& $max$ $F_{\beta}$ $\uparrow$& $max$ $E_{\xi}$ $\uparrow$ \\\hline
		\multirow{5}{*}{Handcrafted}&
		MB~\cite{ZhuLGWW17MB}&.607&.577&.691\\
		&LHM~\cite{peng2014LHM}&.568&.659&.767\\
		&DESM~\cite{cheng2014DESM}&.659&.668&.733\\
		&DCMC~\cite{cong2016DCMC}&.499&.406&.712\\
		&CDCP~\cite{zhu2017CDCP}&.687&.633&.794\\\hline
		\multirow{4}{*}{Deep-based}&
		DMRA~\cite{piao2019DMRA}&.888&.883&.927\\
		&A2dele~\cite{piao2020a2dele}&.886&.892&.929\\
		&SSF~\cite{zhang2020select}&.916&.924&.951\\
		&BBS-Net (ours)&\textbf{.920}&\textbf{.927}&\textbf{.955}\\
		\bottomrule
	\end{tabular}
\end{table}

\begin{table}[t!]
	\footnotesize
	\caption{\small Multiple comparisons of BBS-Net and BBS-Net$^\divideontimes$. The efficient version BBS-Net$^\divideontimes$ has only around $50$ percent parameters of BBS-Net.}
	\renewcommand\arraystretch{0.5}
	\setlength{\tabcolsep}{5.0mm}
	\begin{center}		
		\begin{tabular}{c|c|c|c}\toprule
			\#&Parameters (M)&FLOPs (G)&fps \\\hline
			BBS-Net&49.77&31.40&24.32\\
			BBS-Net$^\divideontimes$&25.96&25.26&25.54\\
			\bottomrule
		\end{tabular}
	\end{center}
	\label{tab:speed}
\end{table}

\begin{table*}[t!]
	\centering
	\caption{ Quantitative comparison of models using S-measure ($S_{\alpha}$), max F-measure ($max F_{\beta}$), max E-measure ($max E_{\xi}$) and MAE ($M$) scores on seven public datasets.
		$\uparrow$ ($\downarrow$) denotes that the higher (lower) the score, the better.
		%
		%
		%
		%
		$\divideontimes$ denotes the efficient version of \ourmodel.
		}
		\vspace{-5pt}
	\label{tab:methodcompare}
	\footnotesize
	\renewcommand{\arraystretch}{0.5}
	\setlength\tabcolsep{2.05pt}
	\begin{tabular}{cr|cccccccccc|cccccccc|cc}
		\toprule
		\multicolumn{1}{c}{\multirow{2}{*}{\rotatebox{90}{Data}}} & \multicolumn{1}{c|}{\multirow{2}{*}{Metric}} & \multicolumn{10}{c|}{Hand-crafted-features-Based Models} & \multicolumn{8}{c|}{CNNs-Based Models}&\multicolumn{2}{c}{BBS-Net} \\
		&&LHM &CDB &DESM &GP &CDCP &ACSD &LBE & DCMC &MDSF &SE &DF &AFNet &CTMF &MMCI &PCF &TANet &CPFP &DMRA&Ours&Ours \\
		&&~\cite{peng2014LHM}&~\cite{liang2018CDB}&\cite{cheng2014DESM}&~\cite{ren2015GP}&~\cite{zhu2017CDCP}&~\cite{ju2014ACSD}&~\cite{feng2016LBE}&~\cite{cong2016DCMC}&~\cite{song2017MDSF}&~\cite{guo2016SE}&~\cite{qu2017DF}&~\cite{wang2019AFNet}&~\cite{han2018CTMF}&~\cite{chen2019MMCI}&~\cite{chen2018PCF}&~\cite{chen2019TANet}&~\cite{zhao2019CPFP}&~\cite{piao2019DMRA}&$\divideontimes$&\\
		\hline\hline
		\multicolumn{1}{c}{\multirow{4}{*}{\rotatebox{90}{NJU2K}}}
		&$S_{\alpha}$ $\uparrow$ &.514&.624&.665&.527&.669&.699&.695&.686&.748&.664&.763&.772&.849&.858&.877&.878&.879&{.886}&.916&\textbf{{.921}}\\
		&$maxF_{\beta}$ $\uparrow$
		&.632&.648&.717&.647&.621&.711&.748&.715&.775&.748&.650&.775&.845&.852&.872&.874&.877&{.886}&.918&\textbf{{.920}}\\
		&$maxE_{\xi}$ $\uparrow$
		&.724&.742&.791&.703&.741&.803&.803&.799&.838&.813&.696&.853&.913&.915&.924&.925&.926&{.927}&.948&\textbf{{.949}}\\
		&$M$ $\downarrow$
		&.205&.203&.283&.211&.180&.202&.153&.172&.157&.169&.141&.100&.085&.079&.059&.060&.053&{.051}&.038&\textbf{.035}\\
		\hline
		\multicolumn{1}{c}{\multirow{4}{*}{\rotatebox{90}{NLPR}}}
		&$S_{\alpha}$ $\uparrow$
		&.630&.629&.572&.654&.727&.673&.762&.724&.805&.756&.802&.799&.860&.856&.874&.886&.888&{.899}&.925&\textbf{{.930}}\\
		&$maxF_{\beta}$ $\uparrow$
		&.622&.618&.640&.611&.645&.607&.745&.648&.793&.713&.778&.771&.825&.815&.841&.863&.867&{.879}&.909&\textbf{{.918}}\\
		&$maxE_{\xi}$ $\uparrow$
		&.766&.791&.805&.723&.820&.780&.855&.793&.885&.847&.880&.879&.929&.913&.925&.941&.932&{.947}&.959&\textbf{{.961}}\\
		&$M$ $\downarrow$
		&.108&.114&.312&.146&.112&.179&.081&.117&.095&.091&.085&.058&.056&.059&.044&.041&.036&{.031}&.026&\textbf{{.023}}\\
		\hline
		\multicolumn{1}{c}{\multirow{4}{*}{\rotatebox{90}{STERE}}}
		&$S_{\alpha}$ $\uparrow$
		&.562&.615&.642&.588&.713&.692&.660&.731&.728&.708&.757&.825&.848&.873&.875&.871&{.879}&.835&.905&\textbf{{.908}}\\
		&$maxF_{\beta}$ $\uparrow$
		&.683&.717&.700&.671&.664&.669&.633&.740&.719&.755&.757&.823&.831&.863&.860&.861&{.874}&.847&.898&\textbf{{.903}}\\
		&$maxE_{\xi}$ $\uparrow$
		&.771&.823&.811&.743&.786&.806&.787&.819&.809&.846&.847&.887&.912&{.927} &.925&.923&.925&.911&.940&\textbf{{.942}}\\
		&$M$ $\downarrow$
		&.172&.166&.295&.182&.149&.200&.250&.148&.176&.143&.141&.075&.086&.068&.064&.060&{.051}&.066&.043&\textbf{{.041}}\\
		\hline
		\multicolumn{1}{c}{\multirow{4}{*}{\rotatebox{90}{DES}}}
		&$S_{\alpha}$ $\uparrow$
		&.578&.645&.622&.636&.709&.728&.703&.707&.741&.741&.752&.770&.863&.848&.842&.858&.872&{.900}&.930&\textbf{{.933}}\\
		&$maxF_{\beta}$ $\uparrow$
		&.511&.723&.765&.597&.631&.756&.788&.666&.746&.741&.766&.728&.844&.822&.804&.827&.846&{.888}&.921&\textbf{{.927}}\\
		&$maxE_{\xi}$ $\uparrow$
		&.653&.830&.868&.670&.811&.850&.890&.773&.851&.856&.870&.881&.932&.928&.893&.910&.923&{.943}&.965&\textbf{{.966}}\\
		&$M$ $\downarrow$
		&.114&.100&.299&.168&.115&.169&.208&.111&.122&.090&.093&.068&.055&.065&.049&.046&.038&{.030}&.022&\textbf{{.021}}\\
		\hline
		\multicolumn{1}{c}{\multirow{4}{*}{\rotatebox{90}{LFSD}}}
		&$S_{\alpha}$ $\uparrow$
		&.553&.515&.716&.635&.712&.727&.729&.753&.694&.692&.783&.738&.788&.787&.786&.801&.828&{.839}&.859&\textbf{{.864}}\\
		&$maxF_{\beta}$ $\uparrow$
		&.708&.677&.762&.783&.702&.763&.722&.817&.779&.786&.813&.744&.787&.771&.775&.796&.826&{.852}&.855&\textbf{{.858}}\\
		&$maxE_{\xi}$ $\uparrow$
		&.763&.871&.811&.824&.780&.829&.797&.856&.819&.832&.857&.815&.857&.839&.827&.847&.863&{.893}&.896&\textbf{{.901}}\\
		&$M$ $\downarrow$
		&.218&.225&.253&.190&.172&.195&.214&.155&.197&.174&.145&.133&.127&.132&.119&.111&.088&{.083}&.076&\textbf{{.072}}\\
		\hline
		\multicolumn{1}{c}{\multirow{4}{*}{\rotatebox{90}{SSD}}}
		&$S_{\alpha}$ $\uparrow$
		&.566&.562&.602&.615&.603&.675&.621&.704&.673&.675&.747&.714&.776&.813&.841&.839&.807&{.857}&.858&\textbf{{.882}}\\
		&$maxF_{\beta}$ $\uparrow$
		&.568&.592&.680&.740&.535&.682&.619&.711&.703&.710&.735&.687&.729&.781&.807&.810&.766&{.844}&.827&\textbf{{.859}}\\
		&$maxE_{\xi}$ $\uparrow$
		&.717&.698&.769&.782&.700&.785&.736&.786&.779&.800&.828&.807&.865&.882&.894&.897&.852&{.906}&{.894}&{\textbf{.919}}\\
		&$M$ $\downarrow$
		&.195&.196&.308&.180&.214&.203&.278&.169&.192&.165&.142&.118&.099&.082&.062&.063&.082&{.058}&.058&\textbf{{.044}}\\
		\hline
		\multicolumn{1}{c}{\multirow{4}{*}{\rotatebox{90}{SIP}}}
		&$S_{\alpha}$ $\uparrow$
		&.511&.557&.616&.588&.595&.732&.727&.683&.717&.628&.653&.720&.716&.833&.842&.835&{.850}&.806&.876&\textbf{{.879}}\\
		&$maxF_{\beta}$ $\uparrow$
		&.574&.620&.669&.687&.505&.763&.751&.618&.698&.661&.657&.712&.694&.818&.838&.830&{.851}&.821&.880&\textbf{{.883}}\\
		&$maxE_{\xi}$ $\uparrow$
		&.716&.737&.770&.768&.721&.838&.853&.743&.798&.771&.759&.819&.829&.897&.901&.895&{.903}&.875&.919&\textbf{{.922}}\\
		&$M$ $\downarrow$
		&.184&.192&.298&.173&.224&.172&.200&.186&.167&.164&.185&.118&.139&.086&.071&.075&{.064}&.085&.056&\textbf{{.055}}\\	
		\bottomrule	
	\end{tabular}
\end{table*}
\myPara{\textbf{Training/Testing.}} We follow the same settings as~\cite{chen2018PCF,piao2019DMRA} for fair comparison. In particular, the training set contains $1,485$ samples from the NJU2K dataset and $700$ samples from the NLPR dataset.
The test set consists of the remaining images from NJU2K ($500$) and NLPR ($300$), and the whole of STERE ($1,000$), DES, LFSD, SSD and SIP.
As for the recent proposde DUT~\cite{piao2019DMRA} dataset, following~\cite{piao2019DMRA}, 
we adopt the same training data of DUT, NJU2K, and NLPR to train the compared deep models (\ie, DMRA~\cite{piao2019DMRA}, A2dele~\cite{piao2020a2dele}, SSF~\cite{zhang2020select}, and our \ourmodel) and test the performance on the test set of DUT. Please refer to \tabref{tab:dut_result} for more details.

\myPara{\textbf{Evaluation Metrics.}}
We employ five widely used metrics, including S-measure ($S_{\alpha}$)~\cite{Fan2017Smeasure},
E-measure ($E_{\xi}$)~\cite{Fan2018Emeasure},
F-measure ($F_{\beta}$)~\cite{Ach2009Fmeasure},
mean absolute error (MAE), and precision-recall (PR) curves to evaluate various methods. 
Evaluation code: \href{http://dpfan.net/d3netbenchmark/}{http://dpfan.net/d3netbenchmark/}.

\subsection{Comparison with SOTAs}\label{sec:compare_sota}
%
\myPara{\textbf{Contenders.}}\label{sec:contenders}
We compare the proposed~\ourmodel~with ten algorithms based on handcrafted features~\cite{cheng2014DESM,cong2016DCMC,feng2016LBE,guo2016SE,ju2014ACSD,liang2018CDB,peng2014LHM,ren2015GP,zhu2017CDCP,song2017MDSF}
and eight methods~\cite{chen2018PCF,chen2019TANet,chen2019MMCI,han2018CTMF,piao2019DMRA,qu2017DF,wang2019AFNet,zhao2019CPFP} that use deep learning.
We train and test these methods using their default settings.
For the methods without released source codes, we compare with their reported results.\par

\myPara{\textbf{Quantitative Results.}} As shown in \tabref{tab:dut_result}, \tabref{tab:methodcompare}, our method outperforms all algorithms based on handcrafted features as well as SOTA CNN-based methods by a large margin, in terms of all four evaluation metrics (\ie, S-measure ($S_{\alpha}$), F-measure ($F_{\beta}$), E-measure ($E_{\xi}$) and MAE ($M$)).
Performance gains over the best compared algorithms (ICCV'19 DMRA~\cite{piao2019DMRA} and CVPR'19 CPFP~\cite{zhao2019CPFP}) are ($2.5\%\sim3.5\%$, $0.7\%\sim3.9\%$, $0.8\%\sim2.3\%$, $0.009\sim0.016$) for the metrics ($S_{\alpha}$, $maxF_{\beta}$, $maxE_{\xi}$, $M$) on seven challenging datasets.
The PR curves of different methods on various datasets are shown in \figref{fig:prcurve}.
It can be easily deduced from the PR curves that our method (\ie, solid red lines) outperforms all the SOTA algorithms.
%
%
%
\par
In terms of speed, \ourmodel~achieves $24.32$ fps on a single GTX 1080Ti GPU (batch size of one), as shown in \tabref{tab:speed}, which is suitable for real-time applications.
In terms of parameters, \ourmodel$^\divideontimes$ contains only around $50$ percent parameters of the \ourmodel~(\ie, $25.96$M vs. $49.77$M), but its performance is similar to the \ourmodel~and also superior to other compared methods (as shown in the last two columns in \tabref{tab:methodcompare}).
It means that \ourmodel$^\divideontimes$ can process more images in the same time (with a larger batch size) for real-world applications.
\par
%
There are three popular backbone models used in deep RGB-D models (\ie, VGG-16~\cite{simonyan2014vgg}, VGG-19~\cite{simonyan2014vgg} and ResNet-50~\cite{He2016resnet}).
To further validate the effectiveness of the proposed method, we provide performance comparisons using different backbones in \tabref{tab:backbone}.
We find that ResNet-50 performs best among the three backbones, and VGG-19 and VGG-16 have similar performances.
Besides, the proposed method exceeds the SOTA methods (\eg, TANet~\cite{chen2019TANet}, CPFP~\cite{zhao2019CPFP}, and DMRA~\cite{piao2019DMRA}) with any of the backbones.

\begin{figure*}[t!]
	\centering
	\begin{overpic}[width=.98\linewidth]{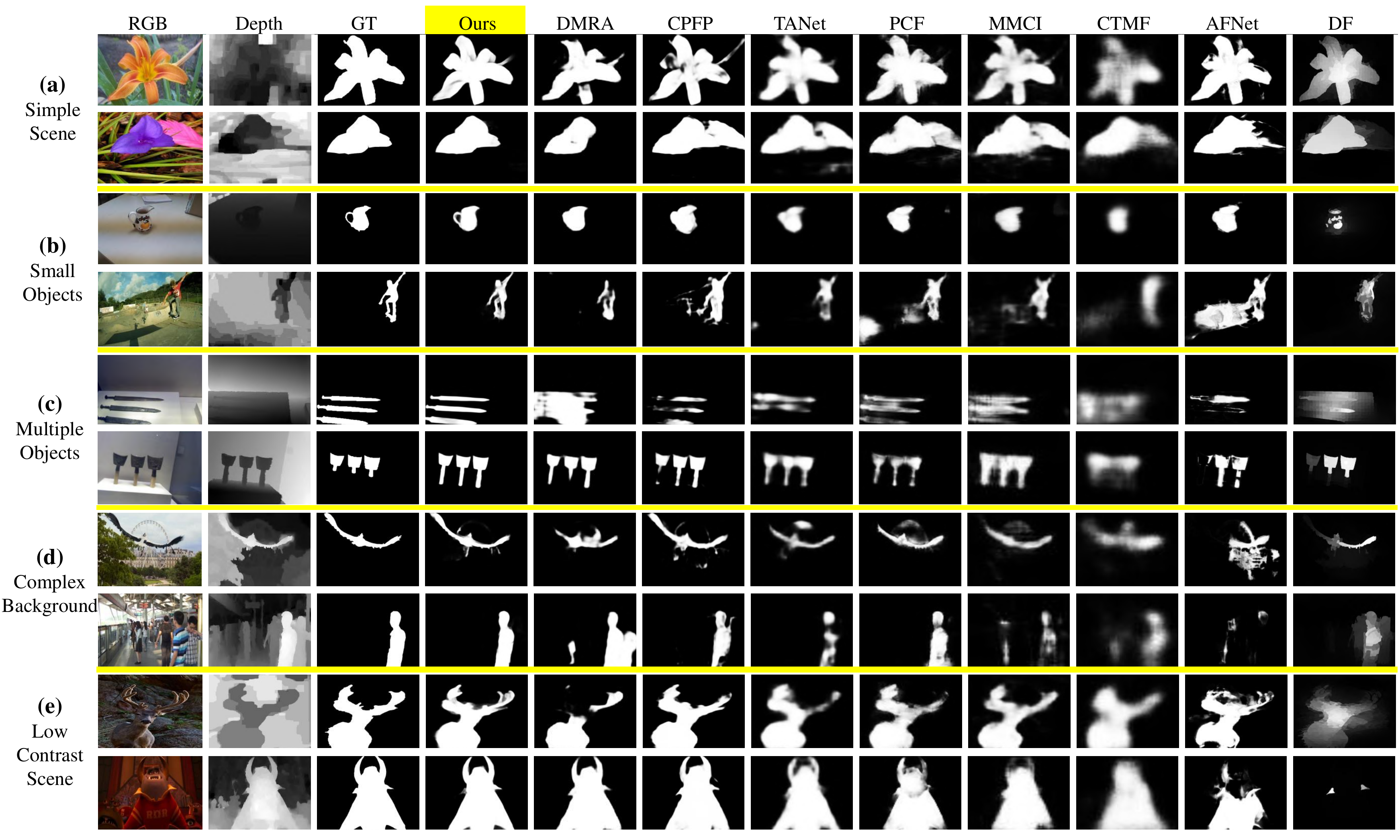}
	\end{overpic}
	\vspace{-10pt}
	\caption{ \textbf{Qualitative visual comparison of our model versus eight SOTA models.} }
	\label{fig:visual_result}
\end{figure*}
\begin{table*}[t!]
	\caption{ Performance comparison using different backbone models. 
		We experiment with multiple popular backbone models used in RGB-D SOD, including VGG-16~\cite{simonyan2014vgg}, VGG-19~\cite{simonyan2014vgg} and ResNet-50~\cite{He2016resnet}.}
		\vspace{-5pt}
	\label{tab:backbone}
	\centering
	\renewcommand{\arraystretch}{0.5}
	\setlength\aboverulesep{0.5pt}\setlength\belowrulesep{1pt}
	\setlength{\tabcolsep}{2.40mm}{
		\footnotesize
		\begin{tabular}{l|cc|cc|cc|cc|cc|cc|cc}
			\toprule
			\multicolumn{1}{c|}{\multirow{2}{*}{{Models}}} & \multicolumn{2}{c|}{NJU2K~\cite{ju2014ACSD}} & \multicolumn{2}{c|}{NLPR~\cite{peng2014LHM}} & \multicolumn{2}{c|}{STERE~\cite{niu2012STERE}}& \multicolumn{2}{c|}{DES~\cite{cheng2014DESM}}& \multicolumn{2}{c|}{LFSD~\cite{li2014LFSD}}& \multicolumn{2}{c|}{SSD~\cite{zhu2017SSD}}&
			\multicolumn{2}{c}{SIP~\cite{fan2019D3Net}} \\
			&$S_{\alpha}\uparrow$&$M\downarrow$&$S_{\alpha}\uparrow$ &$M\downarrow$&$S_{\alpha}\uparrow$ &$M\downarrow$&$S_{\alpha}\uparrow$ &$M\downarrow$&$S_{\alpha}\uparrow$ &$M\downarrow$&$S_{\alpha}\uparrow$ &$M\downarrow$&$S_{\alpha}\uparrow$ &$M\downarrow$\\
			\midrule
			TANet (VGG-16)~\cite{chen2019TANet}&.878&.060 &.886&.041 &.871&.060 &.858&.046 &.801&.111 &.839&.063 &.835&.075\\
			CPFP (VGG-16)~\cite{zhao2019CPFP}&.879&.053 &.888&.036 &.879&.051 &.872&.038 &.828&.088 &.807&.082 &.850&.064\\
			Ours (VGG-16)&.916&.039 &.923&.026 &.896&.046 &.908&.028 &.845&.080 &.858&.055 &.874&.056\\
			\hline
			DMRA (VGG-19)~\cite{piao2019DMRA}&.886&.051 &.899&.031 &.835&.066 &.900&.030 &.839&.083 &.857&.058 &.806&.085\\
			Ours (VGG-19)&.918&.037 &.925&.025 &.901&.043 &.915&.026 &.852&.074 &.855&.056 &.878&\textbf{.054}\\
			\hline
			D3Net (ResNet-50)~\cite{fan2019D3Net}&.900&.041 &.912&.030 &.899&.046 &.898&.031 &.825&.095 &.857&.058 &.860&.063\\
			\rowcolor{mygray}
			Ours (ResNet-50)&\textbf{.921}&\textbf{.035} &\textbf{.930}&\textbf{.023} &\textbf{.908}&\textbf{.041} &\textbf{.933}&\textbf{.021} &\textbf{.864}&\textbf{.072} &\textbf{.882}&\textbf{.044}  &\textbf{.879}&.055\\
			
			\bottomrule
	\end{tabular}}
\end{table*}


\myPara{\textbf{Visual Comparison.}}\label{sec:visual_comparison}
\figref{fig:visual_result} provides examples of maps predicted by our method and several SOTA algorithms.
Visualizations cover simple scenes (a) and various challenging scenarios, including small objects (b), multiple objects (c), complex backgrounds (d), and low contrast scenes (e).\par
First, the first row of (a) shows an easy example. The flower in the foreground is evident in the original RGB image, but the depth map is of low quality and contains some misleading information.
The SOTA algorithms, such as DMRA and CPFP, fail to predict the whole extent of the salient object due to the interference from the depth map. Our method can eliminate the side-effects of the depth map by utilizing the complementary depth information more effectively.
Second, two examples of small objects are shown in (b). Despite the handle of the teapot in the first row being tiny, our method can accurately detect it.
Third, we show two examples with multiple objects in an image in (c).
Our method locates all salient objects in the image. It segments the objects more accurately and generates sharper edges compared to other algorithms.
Even though the depth map in the first row of (c) lacks clear information, our algorithm predicts the salient objects correctly.
Fourth, (d) shows two examples with complex backgrounds.
Here, our method produces reliable results, while other algorithms confuse the background as a salient object.
Finally, (e) presents two examples in which the contrast between the object and the background is low.
Many algorithms fail to detect and segment the entire extent of the salient object.
Our method produces satisfactory results by suppressing background distractors and exploring the informative cues from the depth map.
%

\subsection{Ablation Study}\label{sec:ablation_study}
\begin{figure}[t!]
	\centering
	\begin{overpic}[width=1.0\linewidth]{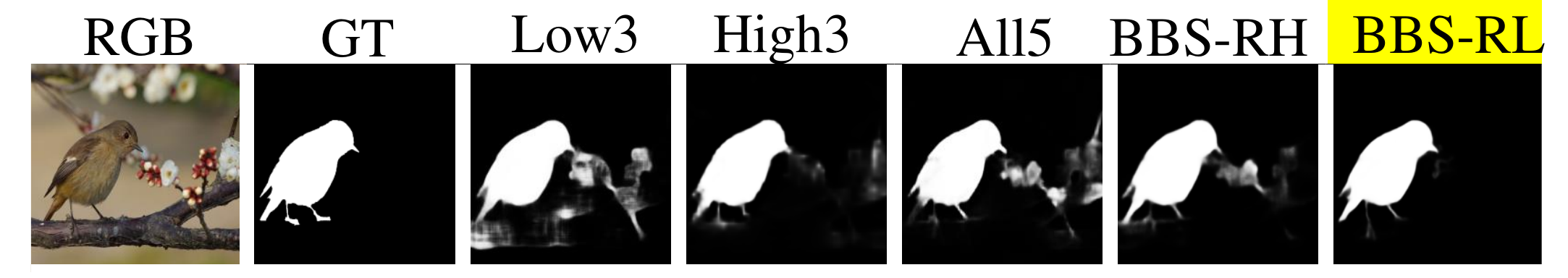}
	\end{overpic}
	\vspace{-20pt}
	\caption{ \textbf{Visual comparison of aggregation strategies.} `Low3' only integrates low-level features (\textit{Conv1}$\sim$\textit{3}), while `High3' aggregates high-level features (\textit{Conv3}$\sim$\textit{5}) for predicting the saliency map. `All5' combines all five-level features directly for prediction. `BBS-RH/BBS-RL' denotes that high-level/low-level features are first refined by the initial map aggregated by the low-level/high-level features and are then integrated to predict the final map.}
	\label{fig:visual_agg}
\end{figure}

\vspace{-5pt}
\myPara{\textbf{Analysis of Different Aggregation Strategies.}}
To validate the effectiveness of our cascaded refinement mechanism,
we conduct several experiments to explore different aggregation strategies. Results are shown in \tabref{tab:agg} and \figref{fig:visual_agg}.
`Low3' means that we only integrate the low-level features (\textit{Conv1}$\sim$\textit{3}) using the decoder without the refinement from the initial map.
Low-level features contain abundant details that are beneficial for refining the object edges, but at the same time introduce a lot of background distraction.
\begin{table*}[t!]
	\caption{Comparison of feature aggregation strategies. 
	1: Only aggregating the low-level features (\textit{Conv1}$\sim$\textit{3}), 2: Only aggregating the high-level features (\textit{Conv3}$\sim$\textit{5}), 3: Directly integrating all five-level features (\textit{Conv1}$\sim$\textit{5}) by a single decoder, 4: Our model without the refinement flow, 5: High-level features (\textit{Conv3}$\sim$\textit{5}) are first refined by the initial map aggregated by low-level features (\textit{Conv1}$\sim$\textit{3}) and are then integrated to generate the final saliency map, and 6: Our cascaded refinement mechanism.}
	\vspace{-5pt}
	\label{tab:agg}
	\centering
	\renewcommand{\arraystretch}{0.5}
	\setlength\aboverulesep{0.5pt}\setlength\belowrulesep{1pt}
	\setlength{\tabcolsep}{2.52mm}{
		\footnotesize
		\begin{tabular}{c|c|cc|cc|cc|cc|cc|cc|cc}
			\toprule
			\multicolumn{1}{c|}{\multirow{2}{*}{{\#}}}&
			\multicolumn{1}{c|}{\multirow{2}{*}{Settings}} & \multicolumn{2}{c|}{NJU2K~\cite{ju2014ACSD}} & \multicolumn{2}{c|}{NLPR~\cite{peng2014LHM}} & \multicolumn{2}{c|}{STERE~\cite{niu2012STERE}}& \multicolumn{2}{c|}{DES~\cite{cheng2014DESM}}& \multicolumn{2}{c|}{LFSD~\cite{li2014LFSD}}& \multicolumn{2}{c|}{SSD~\cite{zhu2017SSD}}&
			\multicolumn{2}{c}{SIP~\cite{fan2019D3Net}} \\
			&&$S_{\alpha}\uparrow$ &$M\downarrow$&$S_{\alpha}\uparrow$ &$M\downarrow$&$S_{\alpha}\uparrow$ &$M\downarrow$&$S_{\alpha}\uparrow$ &$M\downarrow$&$S_{\alpha}\uparrow$ &$M\downarrow$&$S_{\alpha}\uparrow$ &$M\downarrow$&$S_{\alpha}\uparrow$ &$M\downarrow$\\
			\midrule
			1&Low 3 levels&.881&.051 &.882&.038  &.832&.070  &.853&.044  &.779&.110  &.805&.080&.760&.108   \\
			2&High 3 levels&.902&.042
			&.911&.029 &.886&.048 &.912&.026  &.845&.080  &.850&.058 &.833&.073\\
			3&All 5 levels&.905&.042   &.915&.027  &.891&.045 &.901&.028  &.845&.082  &.848&.060&.839&.071   \\
			4&BBS-NoRF&.893&.050 &.904&.035 &.843&.072 &.886&.039 &.804&.105 &.839&.069 &.843&.076\\
			5&BBS-RH &.913&.040  &.922&.028  &.881&.054  &.919&.027  &.833&.085 &.872&.053&.866&.063 \\\hline
			\rowcolor{mygray}
			6&BBS-RL (ours)&\textbf{.921}&\textbf{.035}  &\textbf{.930}&\textbf{.023}  &\textbf{.908}&\textbf{.041} &\textbf{.933}&\textbf{.021}  &\textbf{.864}&\textbf{.072}  &\textbf{.882}&\textbf{.044}&\textbf{.879}&\textbf{.055}  \\
			\bottomrule
	\end{tabular}}
\end{table*}
\begin{table*}[t!]
	\caption{ Ablation analysis of our \ourmodel. `BM' = base model. `CA' = channel attentio. `SA' = spatial attention.  `PTM' = progressively transposed module.}
	\vspace{-5pt}
	\label{tab:ablation}
	\centering
	\renewcommand{\arraystretch}{0.5}
	\setlength\aboverulesep{0.5pt}\setlength\belowrulesep{1pt}
	\footnotesize
	\setlength{\tabcolsep}{2.12mm}{
		\begin{tabular}{c|cccc|cc|cc|cc|cc|cc|cc|cc}
			\toprule
			\multicolumn{1}{c|}{\multirow{2}{*}{{\#}}}&
			\multicolumn{4}{c|}{Settings} &
			\multicolumn{2}{c|}{NJU2K~\cite{ju2014ACSD}} & \multicolumn{2}{c|}{NLPR~\cite{peng2014LHM}} & \multicolumn{2}{c|}{STERE~\cite{niu2012STERE}}& \multicolumn{2}{c|}{DES~\cite{cheng2014DESM}}& \multicolumn{2}{c|}{LFSD~\cite{li2014LFSD}}& \multicolumn{2}{c|}{SSD~\cite{zhu2017SSD}}&
			\multicolumn{2}{c}{SIP~\cite{fan2019D3Net}} \\	&BM&CA&SA&PTM&$S_{\alpha}\uparrow$ &$M\downarrow$&$S_{\alpha}\uparrow$ &$M\downarrow$&$S_{\alpha}\uparrow$ &$M\downarrow$&$S_{\alpha}\uparrow$ &$M\downarrow$&$S_{\alpha}\uparrow$ &$M\downarrow$&$S_{\alpha}\uparrow$ &$M\downarrow$&$S_{\alpha}\uparrow$&$M\downarrow$ \\
			\midrule
			1& \checkmark &          &          &          &.908&.045
			&.918&.029  &.882&.055  &.917&.027  &.842&.083 &.862&.057&.864&.066 \\
			2& \checkmark &\checkmark&          &          &.913&.042
			&.922&.027 &.896&.048 &.923&.025  &.840&.086  &.855&.057&.868&.063\\
			3& \checkmark &          &\checkmark&
			&.912&.045 &.918&.029 &.891&.054 &.914&.029  &.855&.083  &.872&.054&.869&.063 \\
			4&\checkmark&\checkmark&\checkmark&            &.919&.037
			&.928&.026 &.900&.045 &.924&.024  &.861&.074  &.873&.052&.869&.061 \\
			\rowcolor{mygray}\hline
			5& \checkmark &\checkmark&\checkmark&\checkmark&\textbf{.921}&\textbf{.035} &\textbf{.930}&\textbf{.023} &\textbf{.908}&\textbf{.041} &\textbf{.933}&\textbf{.021}  &\textbf{.864}&\textbf{.072}  &\textbf{.882}&\textbf{.044}&\textbf{.879}&\textbf{.055} \\
			\bottomrule
	\end{tabular}}
\end{table*}
Integrating only low-level features produces inadequate results
and generates many distractors
(\eg, the example in \figref{fig:visual_agg}). 
`High3' only integrates the high-level features (\textit{Conv3}$\sim$\textit{5}) to predict the saliency map.
Compared with low-level features, high-level features contain more semantic information.
As a result, they help locate the salient objects and preserve edge information.
Thus, integrating high-level features leads to better results.
`All5' aggregates features from all five levels (\textit{Conv1}$\sim$\textit{5}) directly, using a single decoder for training and testing.
It achieves comparable results with the 'High3' but may include background noise introduced by the low-level features (see column `All5' in \figref{fig:visual_agg}).
`BBS-NoRF' indicates that we directly remove the refinement flow of our model. This leads to poor performance.
`BBS-RH' is a reverse refinement strategy to our cascaded refinement mechanism, where teacher features (\textit{Conv3}$\sim$\textit{5}) are first refined by the initial map aggregated by low-level features (\textit{Conv1}$\sim$\textit{3}) and are then integrated to generate the final saliency map.
It performs worse than the proposed mechanism (BBS-RL), because noise in low-level features cannot be effectively suppressed in this reverse refinement strategy.
Besides, compared to `All5', our method fully utilizes the features at different levels, and thus achieves significant performance improvement (\ie, the last row in \tabref{tab:agg}) with fewer background distractors and sharper edges. 
\par
\begin{figure}[t!]
	\centering
	\begin{overpic}[width=1.0\linewidth]{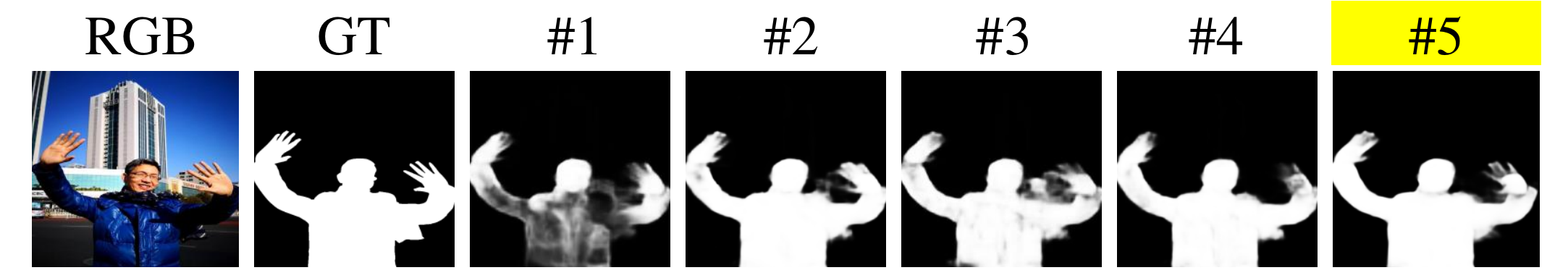}
	\end{overpic}
	\vspace{-20pt}
	\caption{\textbf{Analysis of gradually adding various modules.} The first two columns are the RGB and ground-truth images, respectively. `\#' denotes the corresponding row of \tabref{tab:ablation}. }
	\label{fig:visual_ablation}
\end{figure}
\myPara{\textbf{Impact of Different Modules.}} To validate the effectiveness of the different modules in the proposed \ourmodel, we conduct various experiments, as shown in \tabref{tab:ablation} and \figref{fig:visual_ablation}.
The base model (BM) is our \ourmodel ~without additional modules (\ie, CA, SA, and PTM).
Note that the BM alone performs better than the SOTA methods over almost all datasets, as shown in \tabref{tab:methodcompare} and \tabref{tab:ablation}.
Adding the channel attention (CA) and spatial attention (SA) modules enhances the performance on most of the datasets (see the results shown in the second and third rows of \tabref{tab:ablation}).
When we combine the two modules (the fourth row in \tabref{tab:ablation}), the performance is greatly improved on all datasets, compared to the BM.
We can easily conclude from the `\#2', `\#3' and `\#4' columns in 
\figref{fig:visual_ablation} that the spatial attention and
channel attention mechanisms in DEM allow the model to focus on the informative parts of the depth features, which results in better suppression of background clutter.
Finally, we add a progressively transposed block before the second decoder to gradually upsample the feature map to the same resolution as the ground truth.
The results in the fifth row of \tabref{tab:ablation} and the '\#5' column of \figref{fig:visual_ablation} show that the `PTM' achieves impressive performance gains on all datasets and generates sharper edges with finer details. 
\par
To further analyze the effectiveness of the cascaded decoder, we experiment with changing it to an element-wise summation mechanism.
That is to say, we first change the features from different layers to the same dimension using $1\times1$ convolution and upsampling operation and then fuse them by element-wise summation.
Experimental results in \tabref{tab:cascaded_decoder} show that the cascaded decoder achieves comparable results on SIP, and outperforms the element-wise sum on the other six datasets, which demonstrates its effectiveness.
\begin{table}[t!]
	\vspace{-5pt}
	\caption{  Effectiveness analysis of the cascaded decoder in terms of the S-measure ($S_{\alpha}$) on seven datasets.}
	\vspace{-5pt}
	\label{tab:cascaded_decoder}
	\centering
	\renewcommand{\arraystretch}{0.5}
	\setlength{\tabcolsep}{1.15mm}{
		\footnotesize
		\begin{tabular}{r|ccccccccc}
			\toprule
			{\multirow{2}{*}{Methods}}&NJU2K&NLPR&STERE&DES&SSD&LFSD&SIP \\
			&~\cite{ju2014ACSD}&~\cite{peng2014LHM}&~\cite{niu2012STERE}&~\cite{cheng2014DESM}&~\cite{zhu2017SSD}&~\cite{li2014LFSD}&~\cite{fan2019D3Net}\\\midrule
			Element-wise sum&.915 &.925&.897&.925&.868&.856&\textbf{.880}\\
			\hline
			\rowcolor{mygray}
			Cascaded decoder&\textbf{.921} &\textbf{.930}&\textbf{.908}&\textbf{.933}&\textbf{.882}&\textbf{.864}&.879\\
			\bottomrule
	\end{tabular}}
\end{table}
\begin{table}[t!]
	\caption{\small Hyper-parameter $\alpha$ analysis on the NJU2K dataset. We do not report the result for $\alpha =1$, because its loss of the final predicted map is $0$.}
	\label{tab:hyper}
	\centering
	\renewcommand{\arraystretch}{0.5}
	\setlength\aboverulesep{0.5pt}\setlength\belowrulesep{1pt}
	\setlength{\tabcolsep}{0.98mm}{
		\footnotesize
		\begin{tabular}{c|ccccccccccc}
			\toprule
			$\alpha$&0&0.1&0.2&0.3&0.4&0.5&0.6&0.7&0.8&0.9\\ \hline
			NJU2K ($S_{\alpha}$)&.918&.925&.923&.919&.924&.923&.920&.923&.922&.924\\
			NJU2K (MAE)&.037&.034&.034&.036&.033&.034&.035&.034&.035&.033\\
			\bottomrule
	\end{tabular}}
\end{table}
\par
\myPara{Hyper-parameter Analysis.} We conduct an experiment to discuss the settings of $\alpha$.
As shown in \tabref{tab:hyper}, the performance ($S_{\alpha}$ and MAE) is about the same for different values of $\alpha$, thus we simply set it to $0.5$ to balance the weight between the losses of the initial map and the final map.
\begin{table}[t!]
	\vspace{-5pt}
	\caption{Effectiveness analysis of the depth adapter module in terms of the S-measure ($S_{\alpha}$) on seven datasets. $\divideontimes$ represents the efficient version of \ourmodel, where the two backbones share parameters.}
	\vspace{-5pt}
	\label{tab:dam}
	\centering
	\renewcommand{\arraystretch}{0.5}
	\setlength{\tabcolsep}{0.8mm}{
		\footnotesize
		\begin{tabular}{r|ccccccccc}
			\toprule
			{\multirow{2}{*}{Settings}}&NJU2K&NLPR&STERE&DES&SSD&LFSD&SIP \\
			&~\cite{ju2014ACSD}&~\cite{peng2014LHM}&~\cite{niu2012STERE}&~\cite{cheng2014DESM}&~\cite{zhu2017SSD}&~\cite{li2014LFSD}&~\cite{fan2019D3Net}\\\midrule
			\ourmodel$^\divideontimes$ (w/o DAM)&.905&.922&.899&.928&.856&.841&.849\\
			\hline
			\rowcolor{mygray}
			\ourmodel$^\divideontimes$ (w/ DAM)&\textbf{.916} &\textbf{.925}&\textbf{.905}&\textbf{.930}&\textbf{.858}&\textbf{.859}&\textbf{.876}\\
			\bottomrule
	\end{tabular}}
\end{table}

\par
\myPara{Effectiveness Analysis of the Depth Adapter Module.} To demonstrate the effectiveness of the proposed depth adapter module (DAM), we conduct an experiment in \tabref{tab:dam}.
As shown in the table, \ourmodel$^\divideontimes$ (w/ DAM) performs better than \ourmodel$^\divideontimes$ (w/o DAM) on seven datasets, especially on the dataset of NJU2K, LFSD, and SIP.
The DAM can model the modality difference between the RGB image and depth image, reduces the gap between them. Thus the same backbone is more suitable to extract two different modality features.
\section{Discussion}\label{sec:discussion}
%
%
%

\begin{table}[t!]
	\vspace{-5pt}
	\caption{ S-measure ($S_{\alpha}$) comparison with SOTA RGB SOD methods. `w/o depth' and `w/ depth' represent training and testing the proposed method without/with the depth information (\ie, the inputs of the depth branch are or are not set to zeros).}
	\vspace{-5pt}
	\label{tab:depth_benefits}
	\centering
	\renewcommand{\arraystretch}{0.5}
	\setlength{\tabcolsep}{0.85mm}{
		\footnotesize
		\begin{tabular}{r|ccccccccc}
			\toprule
			\multirow{2}{*}{Methods}&NJU2K&NLPR&STERE&DES&LFSD&SSD&SIP \\
			&~\cite{ju2014ACSD}&~\cite{peng2014LHM}&~\cite{niu2012STERE}&~\cite{cheng2014DESM}&~\cite{li2014LFSD}&~\cite{zhu2017SSD}&~\cite{fan2019D3Net}\\
			\midrule
			PiCANet~\cite{Liu_2018_PiCAN}&.847&.834&.868&.854&.761&.832&-\\
			PAGRN~\cite{Zhang2018PAGR}&.829&.844&.851&.858&.779&.793&-\\
			R3Net~\cite{deng2018r3net}&.837&.798&.855&.847&.797&.815&-\\
			CPD~\cite{Wu2019CPD}&.894&.915&.902&.897&.815&.839&.859\\
			PoolNet~\cite{Liu2019SPBD}&.887&.900&.880&.873&.787&.773&.861\\
			\hline
			BBS-Net (w/o depth)&.914&.925&.915&.912&.836&.855&.875\\
			\rowcolor{mygray}
			BBS-Net (w/ depth) &\textbf{.921}&\textbf{.930}&\textbf{.908}&\textbf{.933}&\textbf{.864}&\textbf{.882}&\textbf{.879}\\
			\bottomrule
	\end{tabular}}
\end{table}

\subsection{Utility of Depth Information}\label{sec:depth_benefits}
To explore whether depth information can really contribute to the performance of SOD, we conduct two experiments, results of which are shown in \tabref{tab:depth_benefits}.
On the one hand, we compare the proposed method with five SOTA RGB SOD methods (\ie, PiCANet~\cite{Liu_2018_PiCAN}, PAGRN~\cite{Zhang2018PAGR}, R3Net~\cite{deng2018r3net}, CPD~\cite{Wu2019CPD}, and PoolNet~\cite{Liu2019SPBD}) by neglecting the depth information.
We train and test CPD and PoolNet using the same training and test sets as our model.
For other methods, we use the published results from \cite{piao2019DMRA}.
It is clear that the proposed methods (\ie, BBS-Net (w/ depth)) can significantly exceed SOTA RGB SOD methods thanks to depth information.
On the other hand, we train and test the proposed method without using the depth information by setting the inputs of the depth branch to zero (\ie, BBSNet (w/o depth)).
Comparing the results of the last two rows in the table, we find that depth information effectively improves the performance of the proposed model (especially over the small datasets, \ie, DES, LFSD, and SSD).\par
\begin{figure}[t!]
	\centering	
	\begin{overpic}[width=1.0\linewidth]{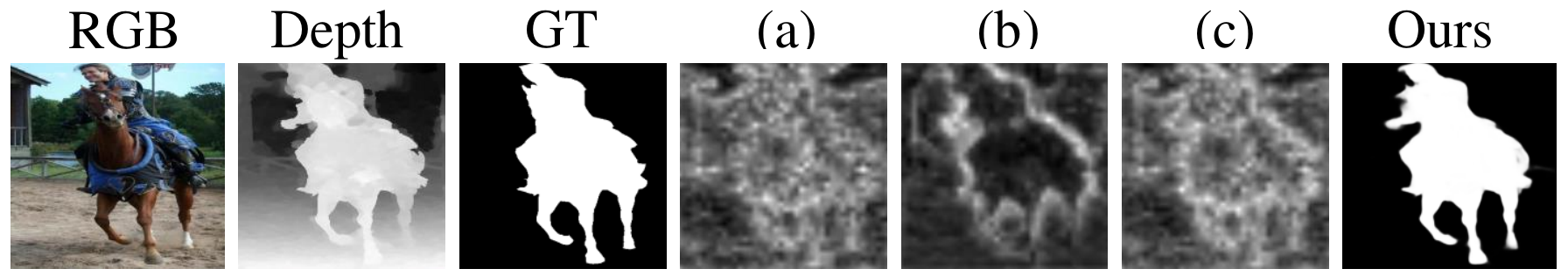}
	\end{overpic}
	\vspace{-20pt}
	\caption{ Feature visualization. Here, (a), (b), and (c) are the average RGB feature, depth feature and cross-modal feature of the \textit{Conv3} layer. To visualize them, we average the
		feature maps along their channel axis to obtain the visualization map. `Ours' refers to the BBS-Net (w/ depth).}
	\label{fig:feature}
\end{figure}

The two experiments together demonstrate the benefits of the depth information for SOD.
Depth map serves as prior knowledge and provides spatial distance information and contour guidance to detect salient objects.
For example, in \figref{fig:feature}, depth feature (b) has high activation on the object border.
Thus, cross-modal feature (c) has clearer borders compared with the original RGB feature (a).

\subsection{Analysis of Post-processing\label{post} Methods}\label{sec:post_process}
\begin{figure}[t!]
	\centering
	\begin{overpic}[width=1.0\linewidth]{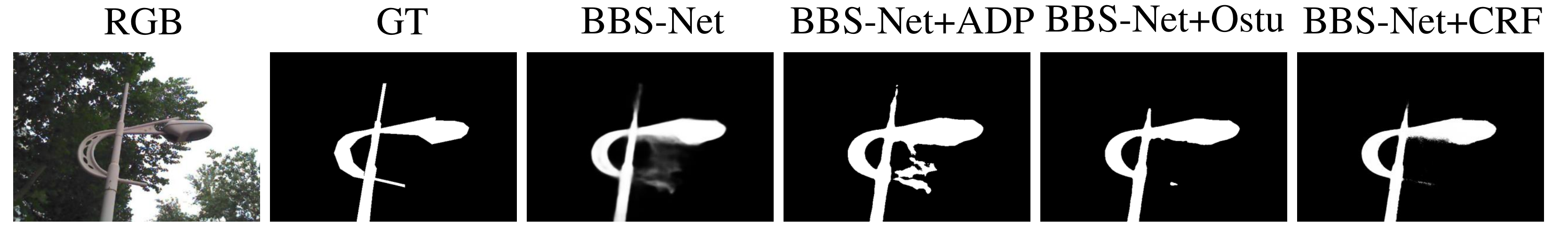}
	\end{overpic}
	\vspace{-20pt}
	\caption{ Visual effects of different post-processing methods. We explore three methods, including the adaptive threshold cut (`ADP' in the paper), Ostu's method and the popular algorithm of conditional random fields (CRF). }
	\label{fig:post_pro}
\end{figure}

According to~\cite{yang2016top,wang2019pyramid,zeng2019joint}, the predicted saliency maps can be further refined by post-processing methods.
This may be useful to sharpen the salient edges and suppress the background response.
We conduct several experiments to study the effects of various post-processing methods, including the adaptive threshold cut (\ie, the threshold is defined as the double of the mean value of the saliency map), Ostu's method~\cite{otsu1979threshold}, and conditional random field (CRF)~\cite{KrahenbuhlK11crf}.
The performance comparisons of the post-processing methods in terms of MAE are shown in \tabref{tab:postpro}, while a visual comparison is provided in \figref{fig:post_pro}.\par

From the results, we draw the following conclusions.
First, the three post-processing methods all make the salient edges sharper, as shown in the fourth to sixth columns in \figref{fig:post_pro}.
%
%
%
\begin{table}[t!]
	\caption{Performance comparison (MAE) of different post-processing strategies on seven datasets. The last column is the time for the post-processing methods to optimize each image. See \secref{post} for details.
	}
	\vspace{-5pt}
	\label{tab:postpro}
	\centering
	\renewcommand{\arraystretch}{0.5}
	\setlength\aboverulesep{0.5pt}\setlength\belowrulesep{1pt}
	\setlength{\tabcolsep}{0.85mm}{
		\footnotesize
		\begin{tabular}{l|cccccccccc}
			\toprule
			\multirow{2}{*}{Strategy}&NJU2K&NLPR&STERE&DES&LFSD&SSD&SIP&time \\
			&~\cite{ju2014ACSD}&~\cite{peng2014LHM}&~\cite{niu2012STERE}&~\cite{cheng2014DESM}&~\cite{li2014LFSD}&~\cite{zhu2017SSD}&~\cite{fan2019D3Net}&ms\\
			\midrule
			BBS-Net&.035&.023&.041&.021&.072&.044&.055&-\\
			BBS-Net+ADP&.050&.024&.049&\textbf{.018}&.072&.053&.055&1.46\\
			BBS-Net+Ostu&\textbf{.030}&\textbf{.020}&.036&\textbf{.018}&.066&.039&\textbf{.051}&0.99\\\rowcolor{mygray}
			\hline
			BBS-Net+CRF&\textbf{.030}&\textbf{.020}&\textbf{.035}&.019&\textbf{.065}&\textbf{.038}&\textbf{.051}&450.8\\
			\bottomrule
			
	\end{tabular}}
	\vspace{5pt}
\end{table}
Second, both Ostu and CRF help reduce the MAE effectively, as shown in \tabref{tab:postpro}.
This is possibly because they can suppress the background noise.
As shown in \figref{fig:post_pro}, Ostu and CRF can significantly reduce the background noise, while the adaptive threshold operation further expands the background blur from the original results of~\ourmodel.
%
%
Further, in terms of overall results, CRF performs the best, while the adaptive threshold algorithm is the worst.
Ostu performs worse than CRF, because it cannot always fully eliminate the background noise (\eg, the fifth and sixth columns in \figref{fig:post_pro}).
\begin{figure}[t!]
	\centering	
	\begin{overpic}[width=1.0\linewidth]{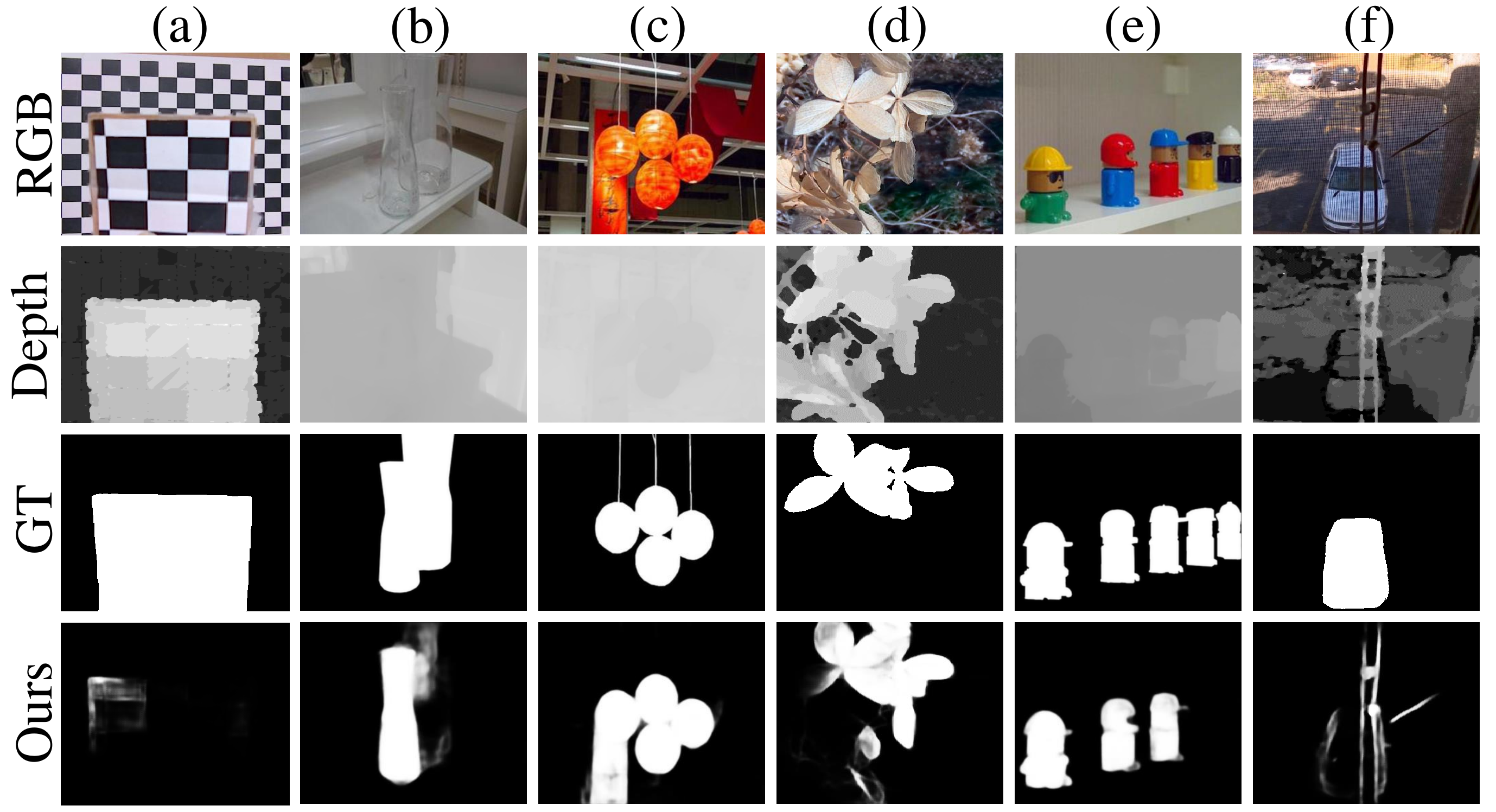}
	\end{overpic}
	\caption{\small Some representative failure cases of the model.} 
	\label{fig:failure_case}
\end{figure}
\subsection{Failure Case Analysis}\label{sec:failure_cases}
We illustrate six representative failure cases in \figref{fig:failure_case}.
The failure examples are divided into four categories.
In the first category, the model either misses the salient object or detects it imperfectly.
For example, in column (a), our model fails to detect the salient object even when the depth map has clear boundaries.
This is because the salient object has the same texture and content layout as the background in the RGB image. Thus, the model cannot find the salient object based only on the borders.
In column (b), our method cannot fully segment the transparent salient objects, since the background has low contrast, and the depth map lacks useful information.
The second situation is that the model identifies the background as the salient part.
For example, the lanterns in column (c) have a similar color to the background wallpaper, which confuses the model into thinking that the wallpaper is the salient object. 
Besides, the background of the RGB image in column (d) is complex and thus our model does not detect the complete salient objects.
The third type of failure case is when an image contains several separate salient objects. In this case, our model may not detect them all.
As shown in column (e), with five salient objects in the RGB images, the model fails to detect the two objects that are far from the camera.
This may be because the model tends to consider the objects that are closer to the camera more salient.
The final case is when salient objects are occluded by non-salient ones.
Note that in column (f), the car is occluded by two ropes in front of the camera. Here our model predicts the ropes as salient objects.\par
Most of these failure cases can be attributed to interference information from the background (\eg, color, contrast, and content).
We propose some ideas that may be useful for solving these failure cases.
The first is to introduce some human-designed prior knowledge, such as providing a boundary that can approximately distinguish the foreground from the background.
Leveraging such prior knowledge, the model may better capture the characteristics of the background and salient objects.
This strategy may contribute significantly to solving the failure cases especially for columns (a) and (b).
Besides, the depth map can also be seen as a type of prior knowledge for this task.
Thus, some failure cases (\ie, (b), (c), and (e)) may be solved when a high-quality depth map is available.
Second, we find that in the current RGB-D datasets, the image pairs for challenging scenarios (\eg, complex backgrounds, low-contrast backgrounds, transparent objects, multiple objects, shielded objects, and small objects) constitute a small fraction of the whole dataset.
Therefore, adding more difficult examples to the training data could help mitigate the failure cases.
Finally, depth maps may sometimes introduce misleading information, such as in column (d).
Considering how to exploit salient cues from the RGB image to suppress the noise in the depth map could be a promising solution.

\subsection{Cross-Dataset Generalization Analysis}\label{sec:cross_datasets}
\begin{table*}[t!]
	\caption{ Performance comparison when training with different datasets. The number in parentheses denotes the number of the corresponding training and test images. See \secref{sec:cross_datasets} for details.
	}
	\vspace{-5pt}
	\label{tab:single_dataset}
	\centering
	\renewcommand{\arraystretch}{0.5}
	\setlength\aboverulesep{0.5pt}\setlength\belowrulesep{1pt}
	\setlength{\tabcolsep}{1.80mm}{
		\footnotesize
		\begin{tabular}{r|cc|cc|cc|cc|cc||cc|cc|cc}
			\toprule
			\multicolumn{1}{c|}{\multirow{2}{*}{{\diagbox{Train}{Test}}}} & \multicolumn{2}{c|}{NJU2K~(1285)} & \multicolumn{2}{c|}{NLPR~(300)} & \multicolumn{2}{c|}{STERE~(300)}&
			\multicolumn{2}{c|}{SIP~(229)}&
			\multicolumn{2}{c||}{DUT~(500)}& \multicolumn{2}{c|}{Self}& \multicolumn{2}{c|}{Mean Others}& \multicolumn{2}{c}{Drop~$\downarrow$} \\
			&$S_{\alpha}\uparrow$&$F_{\beta}$ $\uparrow$&$S_{\alpha}\uparrow$&$F_{\beta}$ $\uparrow$ &$S_{\alpha}\uparrow$&$F_{\beta}$ $\uparrow$ &$S_{\alpha}\uparrow$&$F_{\beta}$ $\uparrow$ &$S_{\alpha}\uparrow$ &$F_{\beta}$ $\uparrow$&$S_{\alpha}\uparrow$&$F_{\beta}$ $\uparrow$ &$S_{\alpha}\uparrow$&$F_{\beta}$ $\uparrow$ &$S_{\alpha}$&$F_{\beta}$\\\midrule
			NJU2K~(700)&\textbf{.902}&\textbf{.894} &.834&.795 &.864&.846 &.802&.782 &.741&.691 &.902&.894 &.810&.779 &10.2\%&12.9\%\\
			NLPR~(700)&.712&.689 &\textbf{.919}&\textbf{.903} &.876&.882 &.883&.881 &.795&.779 &.919&.903 &.817&.808 &11.2\%&10.5\% \\
			STERE~(700)&.779&.741 &.897&.868 &\textbf{.915}&\textbf{.913} &.900&.900 &.724&.731 &.915&.913 &.825&.810 &9.8\%&11.3\%  \\
			SIP~(700)&.436&.325 &.618&.528 &.534&.479 &\textbf{.963}&\textbf{.972} &.423&.303 &.963&.972 &.503&.409 &47.8\%&57.9\%  \\
			DUT~(700)&.751&.777 &.808&.761 &.736&.764 &.801&.802 &\textbf{.887}&\textbf{.877} &.887&.877 &.774&.776 &12.7\%&11.5\%\\\hline
			Mean Others&.670&.633 &.789&.738 &.753&.743 &.847&.841 &.671&.626 &-&- &-&- &-&-\\
			
			\bottomrule
	\end{tabular}}
\end{table*}

\begin{table*}[t!]
	\caption{Performance comparison when training with different combinations of multiple datasets. `NJ', `NL', `ST', `SI' and `DU' represent NJU2K, NLPR, STERE, SIP and DUT, respectively. The number in parentheses denotes the number of corresponding training and test images. The number of training images for each dataset is $700$. 
	The training and test sets will be available at: \supp{\href{ https://drive.google.com/drive/folders/1UYGyg50-0y7i21tWREVMcBpatQPFpSr0?usp=sharing} https://drive.google.com/drive/folders/1UYGyg50-0y7i21tWREVMcBpatQPFpSr0?usp=sharing.}}
	\vspace{-5pt}
	\label{tab:multi_dataset}
	\centering
	\renewcommand{\arraystretch}{0.5}
	\setlength\aboverulesep{0.5pt}\setlength\belowrulesep{1pt}
	\setlength{\tabcolsep}{0.18mm}{
		\footnotesize
		\begin{tabular}{r|ccc|ccc|ccc|ccc|ccc|ccc|ccc|ccc}
			\toprule
			\multicolumn{1}{c|}{\multirow{2}{*}{{\diagbox{Train}{Test}}}} & \multicolumn{3}{c|}{NJ~(1285)} & \multicolumn{3}{c|}{NL~(300)} & \multicolumn{3}{c|}{ST~(300)}& \multicolumn{3}{c|}{DES~(135)}& \multicolumn{3}{c|}{LFSD~(80)}& \multicolumn{3}{c|}{SSD~(80)}&
			\multicolumn{3}{c|}{SI~(229)}&\multicolumn{3}{c}{DU~(500)} \\
			&$S_{\alpha}\uparrow$&$F_{\beta}$ $\uparrow$&$M\downarrow$&$S_{\alpha}\uparrow$&$F_{\beta}$ $\uparrow$ &$M\downarrow$&$S_{\alpha}\uparrow$&$F_{\beta}$ $\uparrow$ &$M\downarrow$&$S_{\alpha}\uparrow$&$F_{\beta}$ $\uparrow$ &$M\downarrow$&$S_{\alpha}\uparrow$ &$F_{\beta}$ $\uparrow$&$M\downarrow$&$S_{\alpha}\uparrow$&$F_{\beta}$ $\uparrow$ &$M\downarrow$&$S_{\alpha}\uparrow$&$F_{\beta}$ $\uparrow$ &$M\downarrow$&$S_{\alpha}\uparrow$&$F_{\beta}$ $\uparrow$ &$M\downarrow$\\\midrule
			NJ+NL~(1,400)&.911&.905&.039 &{.926}&{.916}&{.025} &.899&.898&.044 &.934&.932&.019 &.865&.862&.070 &.861&{.836}&.054 &.890&.893&.048&.799&.748&.095\\
			NJ+ST~(1,400)&.913&.909&{.038} &.885&.859&.040 &.916&.912&{.035} &.927&.910&.022 &.853&.836&.078 &{.869}&.848&{.054} &.885&.882&.052&.729&.719&.118\\
			NJ+DU~(1,400)&.906&.893&.043 &.852&.802&.050 &.875&.854&.053 &.884&.859&.037 &.861&.854&.070 &.862&.839&.053 &.834&.825&.075 &.905&.903&.041 \\
			NL+ST~(1,400)&.781&.748&.097 &.930&.919&.024 &.919&.920&.032 &{.942}&.938&.019 &.672&.645&.161 &.774&.722&.090 &.895&.894&.047 &.836&.819&.070\\
			NL+DU~(1,400)&.777&.771&.104 &.923&.908&.023 &.878&.882&.052 &.940&.936&.019 &.717&.728&.135 &.801&.774&.091 &.886&.888&.054 &.905&.903&.040\\
			ST+DU~(1,400)&.821&.794&.082 &.893&.863&.036 &.917&.914&.034 &.940&.935&.020 &.762&.734&.125 &.777&.736&.092 &{.907}&{.910}&{.039} &.913&.914&.037\\
			NJ+NL+ST~(2,100)&.913&.910&.038 &.923&.904&.027 &{.922}&{.924}&{.033} &.943&.939&.018 &.865&.858&.072 &.853&.818&.056 &.902&.905&.043 &.816&.780&.088\\
			NJ+NL+DU~(2,100)&.911&.905&.041 &.924&.909&.027 &.902&.901&.043 &.942&.939&.018 &{.865}&.856&{.067} &.866&.838&.051 &.894&.897&.048 &.916&.915&.036\\
			NJ+ST+DU~(2,100)&{.910}&.903&.041 &.890&.867&.039 &.923&.923&.031 &.932&.918&.021 &.859&.851&.073 &.863&.838&.055 &.896&.899&.046 &{.917}&{.916}&{.035}\\
			NL+ST+DU~(2,100)&.825&.808&.079 &.924&.911&{.026} &.919&.920&.033 &.946&.944&.017 &.751&.732&.125 &.797&.758&.082 &{.901}&.905&{.043} &.916&.911&.036\\
			\hline
			NJ+NL+ST+DU~(2,800)&{.912}&{.905}&{.039} &.932&.917&.024 &.921&.920&{.033} &.946&{.942}&{.018} &{.864}&{.856}&{.070} &.858&.829&{.054} &.903&.905&{.042} &.917&.913&.037\\
			
			\bottomrule
	\end{tabular}}
\end{table*}

For a deep model to obtain reasonable performance in real-world scenarios, it not only requires an efficient design but must also be trained on a high-quality dataset with a great generalization power.
A good dataset usually contains sufficient images, with all types of variations that occur in reality, so that deep models trained on it can generalize well to the real world. 
In the area of RGB-D SOD, there are several large-scale datasets (\ie, NJU2K, NLPR, STERE, SIP, and DUT), with around $1,000$ training images. \par

\myPara{\textbf{Single Dataset Generalization Analysis.}} Here, we conduct cross-dataset generalization experiments on the above-mentioned five datasets to measure their generalization ability.
%
%
To make fair comparisons among multiple datasets, we balance the datasets with equal number of training samples.
Specifically, we randomly choose $700$ image pairs in each dataset for training, and the remaining images are used for testing.
We then retrain the proposed model on a single training set, and test it on all four test sets.
The results are summarized in \tabref{tab:single_dataset}.
`Self' represents the results of training and testing on the same dataset.
`Mean Others' indicates the average performance on all test sets except ‘self’.
`Drop' means the (percent) drop from `Self' to `Mean Others'.
%
%
%
First, it can be seen from the table that NJU2K and DUT are the hardest datasets since their `Mean Others' of column `NJU2K' and `DUT' are significantly lower than the other three datasets.
This may be because the two datasets include multiple challenging scenes (e.g., transparent objects, multiple objects, complex backgrounds, \etc).
Second, STERE has the best generalization ability, because the average drop of $S_{\alpha}$ and $F_{\beta}$ is lowest among all five datasets.
Besides, SIP generalizes worst (\ie, the drop is the largest among all five datasets), since it mainly focuses on a single person or multiple persons in the wild.
We also notice that the score of the SIP column (`Mean Others') is the highest.
This is likely because the quality of the depth maps captured by the Huawei Mate10 is higher than that produced by traditional devices.
Finally, none of the models trained with a single dataset perform best over all test sets.
Thus, we further explore training on different combinations of datasets with the aim of building a dataset with a strong generalization ability for future research.\par

\myPara{\textbf{Dataset Combination for Generalization Improvement.}} 
According to the results in \tabref{tab:single_dataset}, the model trained on the SIP dataset does not generalize well to other datasets, so we discard it. 
We thus select four relatively large-scale datasets, \ie, NJU2K, NLPR, STERE, and DUT, to conduct our multi-dataset training experiments.
As shown in \tabref{tab:multi_dataset}, we consider all possible training combinations of these four datasets and test the models on all available test sets.
From the results in the table, we draw the following conclusions.
%
%
%
%
First, more training examples do not necessarily lead to better performance on some test sets.
For example, although `NJ+NL+ST', `NJ+NL+DU' and `NJ+NL+ST+DU' contain external training sets, unlike `NJ+NL', they perform similarly with `NJ+NL' on the test set of `NL'.
%
%
Second, including the NJU2K dataset is important for the model to generalize well to small datasets (\ie, LFSD, SSD).
The model trained using the combinations without NJU2K (\ie, `NL+ST' `NL+DU', `ST+DU' and `NL+ST+DU') all obtain low F-measure values (less than $0.8$) on the LFSD and SSD test sets.
In contrast, including `NJ' in the training sets increases the F-measures on the LFSD and SSD datasets by over $0.05$.
Finally, including more examples in the training sets can improve the stability of the model, as it allows diverse scenarios to be taken into consideration.
Thus, the model trained on `NJ+NL+ST+DU', which has the most examples, obtains the best, or are very close to the best, performance.
Due to the limited size of current RGB-D datasets, it is hard for a model trained using a single dataset to perform well under various scenarios.
Thus, we recommend training a model using a combination of datasets with diverse examples to avoid model over-fitting issues.
To promote the development of RGB-D SOD, we hope more challenging RGB-D datasets with diverse examples and high-quality depth maps can be proposed in the future.
%

\section{Conclusion}
In this paper, we present a Bifurcated Backbone Strategy Network (\ourmodel)~for the RGB-D SOD.
To effectively suppress the intrinsic distractors in low-level cross-modal features, we propose to leverage the characteristics of multi-level cross-modal features in a cascaded refinement way: \emph{low-level features are refined by the initial saliency map that is produced by the high-level cross-modal features.}
Besides, we introduce a depth-enhanced module to excavate the informative cues from the depth features in the channel and spatial views, in order to improve the cross-modal compatibility when merging RGB and depth features.
Experiments on eight challenging datasets demonstrate that \ourmodel~outperforms $18$ SOTA models, by a large margin, under multiple evaluation metrics.
%
%
Finally, we conduct a comprehensive analysis of the existing RGB-D datasets and introduce a powerful training set with a strong generalization ability for future research. 
%

\ifCLASSOPTIONcaptionsoff
  \newpage
\fi



%
%
%
\bibliographystyle{IEEEtran}
\bibliography{BBS-Net}

\begin{thebibliography}{100}
\providecommand{\url}[1]{#1}
\csname url@samestyle\endcsname
\providecommand{\newblock}{\relax}
\providecommand{\bibinfo}[2]{#2}
\providecommand{\BIBentrySTDinterwordspacing}{\spaceskip=0pt\relax}
\providecommand{\BIBentryALTinterwordstretchfactor}{4}
\providecommand{\BIBentryALTinterwordspacing}{\spaceskip=\fontdimen2\font plus
\BIBentryALTinterwordstretchfactor\fontdimen3\font minus
  \fontdimen4\font\relax}
\providecommand{\BIBforeignlanguage}[2]{{%
\expandafter\ifx\csname l@#1\endcsname\relax
\typeout{** WARNING: IEEEtran.bst: No hyphenation pattern has been}%
\typeout{** loaded for the language `#1'. Using the pattern for}%
\typeout{** the default language instead.}%
\else
\language=\csname l@#1\endcsname
\fi
#2}}
\providecommand{\BIBdecl}{\relax}
\BIBdecl

\bibitem{fan2020bbs}
D.-P. Fan, Y.~Zhai, A.~Borji, J.~Yang, and L.~Shao, ``{BBS-Net: RGB-D Salient
  Object Detection with a Bifurcated Backbone Strategy Network},'' in
  \emph{ECCV}, 2020, pp. 275--292.

\bibitem{borji2015salient}
A.~Borji, M.-M. Cheng, H.~Jiang, and J.~Li, ``Salient object detection: A
  benchmark,'' \emph{IEEE TIP}, vol.~24, no.~12, pp. 5706--5722, 2015.

\bibitem{wang2019salient}
W.~Wang, Q.~Lai, H.~Fu, J.~Shen, H.~Ling, and R.~Yang, ``Salient object
  detection in the deep learning era: An in-depth survey,'' \emph{IEEE TPAMI},
  2021.

\bibitem{ChengLLZRT19bing}
M.-M. Cheng, Y.~Liu, W.~Lin, Z.~Zhang, P.~L. Rosin, and P.~H.~S. Torr,
  ``{BING:} binarized normed gradients for objectness estimation at 300fps,''
  \emph{CVM}, vol.~5, no.~1, pp. 3--20, 2019.

\bibitem{cheng2017retrival}
M.-M. Cheng, Q.~Hou, S.~Zhang, and P.~{L.Rosin}, ``Intelligent visual media
  processing: When graphics meets vision,'' \emph{JCST}, vol.~32, no.~1, pp.
  110--121, 2017.

\bibitem{Wang2015sal_seg}
W.~Wang, J.~Shen, R.~Yang, and F.~Porikli, ``Saliency-aware video object
  segmentation,'' \emph{IEEE TPAMI}, vol.~40, no.~1, pp. 20--33, 2017.

\bibitem{Cheng2010RFA}
M.-M. Cheng, F.-L. Zhang, N.~J. Mitra, X.~Huang, and S.-M. Hu, ``Repfinder:
  Finding approximately repeated scene elements for image editing,''
  \emph{TOG}, vol.~29, no.~4, pp. 83:1--83:8, 2010.

\bibitem{Fan2019VSOD}
D.-P. Fan, W.~Wang, M.-M. Cheng, and J.~Shen, ``Shifting more attention to
  video salient object detection,'' in \emph{CVPR}, 2019, pp. 8554--8564.

\bibitem{Yan_2019_video}
P.~Yan, G.~Li, Y.~Xie, Z.~Li, C.~Wang, T.~Chen, and L.~Lin, ``Semi-supervised
  video salient object detection using pseudo-labels,'' in \emph{ICCV}, 2019,
  pp. 7284--7293.

\bibitem{borji2012cvpr}
A.~Borji, S.~Frintrop, D.~{N.Sihite}, and L.~{Itti}, ``Adaptive object tracking
  by learning background context,'' in \emph{CVPRW}, 2012, pp. 23--30.

\bibitem{Hong2015tracking}
S.~Hong, T.~You, S.~Kwak, and B.~Han, ``Online tracking by learning
  discriminative saliency map with convolutional neural network,'' in
  \emph{ICML}, 2015, pp. 597--606.

\bibitem{cheng2015GC}
M.-M. Cheng, N.~J. Mitra, X.~Huang, P.~H.~S. Torr, and S.-M. Hu, ``Global
  contrast based salient region detection,'' \emph{IEEE TPAMI}, vol.~37, no.~3,
  pp. 569--582, 2015.

\bibitem{zhang2016co}
D.~Zhang, D.~Meng, and J.~Han, ``Co-saliency detection via a self-paced
  multiple-instance learning framework,'' \emph{IEEE TPAMI}, vol.~39, no.~5,
  pp. 865--878, 2016.

\bibitem{borji2012state}
A.~Borji and L.~Itti, ``State-of-the-art in visual attention modeling,''
  \emph{IEEE TPAMI}, vol.~35, no.~1, pp. 185--207, 2012.

\bibitem{borji2019saliency}
A.~Borji, ``Saliency prediction in the deep learning era: Successes and
  limitations,'' \emph{IEEE TPAMI}, pp. 679--700, 2019.

\bibitem{Liu2019SPBD}
J.-J. Liu, Q.~Hou, M.-M. Cheng, J.~Feng, and J.~Jiang, ``A simple pooling-based
  design for real-time salient object detection,'' in \emph{CVPR}, 2019, pp.
  3917--3926.

\bibitem{wang2018salient}
L.~Wang, L.~Wang, H.~Lu, P.~Zhang, and X.~Ruan, ``Salient object detection with
  recurrent fully convolutional networks,'' \emph{IEEE TPAMI}, vol.~41, no.~7,
  pp. 1734--1746, 2018.

\bibitem{chen2019TANet}
H.~Chen and Y.~Li, ``Three-stream attention-aware network for {RGB-D} salient
  object detection,'' \emph{IEEE TIP}, vol.~28, no.~6, pp. 2825--2835, 2019.

\bibitem{piao2019DMRA}
Y.~Piao, W.~Ji, J.~Li, M.~Zhang, and H.~Lu, ``Depth-induced multi-scale
  recurrent attention network for saliency detection,'' in \emph{ICCV}, 2019,
  pp. 7254--7263.

\bibitem{li2016saliency}
N.~Li, J.~Ye, Y.~Ji, H.~Ling, and J.~Yu, ``Saliency detection on light field,''
  \emph{IEEE TPAMI}, vol.~39, no.~8, pp. 1605--1616, 2016.

\bibitem{zhao2019CPFP}
J.-X. Zhao, Y.~Cao, D.-P. Fan, M.-M. Cheng, X.-Y. Li, and L.~Zhang, ``Contrast
  prior and fluid pyramid integration for {RGBD} salient object detection,'' in
  \emph{CVPR}, 2019, pp. 3927--3936.

\bibitem{chen2018PCF}
H.~Chen and Y.~Li, ``Progressively complementarity-aware fusion network for
  {RGB-D} salient object detection,'' in \emph{CVPR}, 2018, pp. 3051--3060.

\bibitem{chen2019MMCI}
H.~Chen, Y.~Li, and D.~Su, ``Multi-modal fusion network with multi-scale
  multi-path and cross-modal interactions for {RGB-D} salient object
  detection,'' \emph{IEEE TCYBERNETICS}, vol.~86, pp. 376--385, 2019.

\bibitem{li2020crossmodal}
G.~Li, Z.~Liu, L.~Ye, Y.~Wang, and H.~Ling, ``{Cross-Modal Weighting Network
  for {RGB-D} Salient Object Detection},'' in \emph{ECCV}, 2020, pp. 665--681.

\bibitem{guo2016SE}
J.~Guo, T.~Ren, and J.~Bei, ``Salient object detection for {RGB-D} image via
  saliency evolution,'' in \emph{IEEE ICME}, 2016, pp. 1--6.

\bibitem{feng2016LBE}
D.~Feng, N.~Barnes, S.~You, and C.~McCarthy, ``Local background enclosure for
  {RGB-D} salient object detection,'' in \emph{CVPR}, 2016, pp. 2343--2350.

\bibitem{LIU2019SSRC}
Z.~Liu, S.~Shi, Q.~Duan, W.~Zhang, and P.~Zhao, ``Salient object detection for
  {RGB-D} image by single stream recurrent convolution neural network,''
  \emph{Neurocomputing}, vol. 363, pp. 46--57, 2019.

\bibitem{zhu2019PDNet}
C.~{Zhu}, X.~{Cai}, K.~{Huang}, T.~H. {Li}, and G.~{Li}, ``Pdnet: Prior-model
  guided depth-enhanced network for salient object detection,'' in \emph{IEEE
  ICME}, 2019, pp. 199--204.

\bibitem{Wu2019CPD}
Z.~Wu, L.~Su, and Q.~Huang, ``Cascaded partial decoder for fast and accurate
  salient object detection,'' in \emph{CVPR}, 2019, pp. 3907--3916.

\bibitem{wang2019AFNet}
N.~Wang and X.~Gong, ``Adaptive fusion for {RGB-D} salient object detection,''
  \emph{IEEE Access}, vol.~7, pp. 55\,277--55\,284, 2019.

\bibitem{cong2019HSCS}
R.~{Cong}, J.~Lei, H.~Fu, Q.~Huang, X.~Cao, and N.~Ling, ``{HSCS}: Hierarchical
  sparsity based co-saliency detection for {RGBD} images,'' \emph{IEEE TMM},
  vol.~21, no.~7, pp. 1660--1671, 2019.

\bibitem{peng2014LHM}
H.~Peng, B.~Li, W.~Xiong, W.~Hu, and R.~Ji, ``{RGBD} salient object detection:
  a benchmark and algorithms,'' in \emph{ECCV}, 2014, pp. 92--109.

\bibitem{fan2014DSP}
X.~Fan, Z.~Liu, and G.~Sun, ``Salient region detection for stereoscopic
  images,'' in \emph{DSP}, 2014, pp. 454--458.

\bibitem{fang2014TIP}
Y.~Fang, J.~Wang, M.~{Narwaria}, P.~{Le Callet}, and W.~Lin, ``Saliency
  detection for stereoscopic images,'' \emph{IEEE TIP}, vol.~23, no.~6, pp.
  2625--2636, 2014.

\bibitem{cheng2014DESM}
Y.~Cheng, H.~Fu, X.~Wei, J.~Xiao, and X.~Cao, ``Depth enhanced saliency
  detection method,'' in \emph{ICIMCS}, 2014, pp. 23--27.

\bibitem{zhu2017CDCP}
C.~Zhu, G.~Li, W.~Wang, and R.~Wang, ``An innovative salient object detection
  using center-dark channel prior,'' in \emph{CVPRW}, 2017, pp. 1509--1515.

\bibitem{fan2019D3Net}
D.-P. Fan, Z.~Lin, Z.~Zhang, M.~Zhu, and M.-M. Cheng, ``{Rethinking RGB-D
  Salient Object Detection: Models, Data Sets, and Large-Scale Benchmarks},''
  \emph{IEEE TNNLS}, pp. 2075--2089, 2020.

\bibitem{ju2014ACSD}
R.~Ju, L.~Ge, W.~Geng, T.~Ren, and G.~Wu, ``Depth saliency based on anisotropic
  center-surround difference,'' in \emph{ICIP}, 2014, pp. 1115--1119.

\bibitem{niu2012STERE}
Y.~Niu, Y.~Geng, X.~Li, and F.~Liu, ``Leveraging stereopsis for saliency
  analysis,'' in \emph{CVPR}, 2012, pp. 454--461.

\bibitem{liu2010learning}
T.~Liu, Z.~Yuan, J.~Sun, J.~Wang, N.~Zheng, X.~Tang, and H.-Y. Shum, ``Learning
  to detect a salient object,'' \emph{IEEE TPAMI}, vol.~33, no.~2, pp.
  353--367, 2010.

\bibitem{achanta2009frequency}
R.~Achanta, S.~Hemami, F.~Estrada, and S.~Susstrunk, ``Frequency-tuned salient
  region detection,'' in \emph{CVPR}, 2009, pp. 1597--1604.

\bibitem{fan2018foreground}
D.-P. Fan, M.-M. Cheng, J.-J. Liu, S.-H. Gao, Q.~Hou, and A.~Borji, ``Salient
  objects in clutter: Bringing salient object detection to the foreground,'' in
  \emph{ECCV}, 2018, pp. 186--202.

\bibitem{LiY16}
G.~Li and Y.~Yu, ``Deep contrast learning for salient object detection,'' in
  \emph{CVPR}, 2016, pp. 478--487.

\bibitem{ZhangWLWY17}
P.~Zhang, D.~Wang, H.~Lu, H.~Wang, and B.~Yin, ``Learning uncertain
  convolutional features for accurate saliency detection,'' in \emph{ICCV},
  2017, pp. 212--221.

\bibitem{zhao2020suppress}
X.~Zhao, Y.~Pang, L.~Zhang, H.~Lu, and L.~Zhang, ``Suppress and balance: A
  simple gated network for salient object detection,'' in \emph{ECCV}, 2020,
  pp. 35--51.

\bibitem{itti1998model}
L.~Itti, C.~Koch, and E.~Niebur, ``A model of saliency-based visual attention
  for rapid scene analysis,'' \emph{IEEE TPAMI}, vol.~20, no.~11, pp.
  1254--1259, 1998.

\bibitem{li2015visual}
G.~Li and Y.~Yu, ``Visual saliency based on multiscale deep features,'' in
  \emph{CVPR}, 2015, pp. 5455--5463.

\bibitem{cheng2013efficient}
M.-M. Cheng, J.~Warrell, W.-Y. Lin, S.~Zheng, V.~Vineet, and N.~Crook,
  ``Efficient salient region detection with soft image abstraction,'' in
  \emph{ICCV}, 2013, pp. 1529--1536.

\bibitem{Chen_2018_reverse}
S.~Chen, X.~Tan, B.~Wang, H.~Lu, X.~Hu, and Y.~Fu, ``Reverse attention-based
  residual network for salient object detection,'' \emph{IEEE TIP}, vol.~29,
  pp. 3763--3776, 2020.

\bibitem{Zhang2018PAGR}
X.~Zhang, T.~Wang, J.~Qi, H.~Lu, and G.~Wang, ``Progressive attention guided
  recurrent network for salient object detection,'' in \emph{CVPR}, 2018, pp.
  714--722.

\bibitem{Su_2019_ICCV}
J.~Su, J.~Li, Y.~Zhang, C.~Xia, and Y.~Tian, ``Selectivity or invariance:
  Boundary-aware salient object detection,'' in \emph{ICCV}, 2019, pp.
  3799--3808.

\bibitem{zhang2020multi}
L.~Zhang, J.~Wu, T.~Wang, A.~Borji, G.~Wei, and H.~Lu, ``A multistage
  refinement network for salient object detection,'' \emph{IEEE TIP}, vol.~29,
  pp. 3534--3545, 2020.

\bibitem{Li_2019_video}
H.~Li, G.~Chen, G.~Li, and Y.~Yu, ``Motion guided attention for video salient
  object detection,'' in \emph{ICCV}, 2019, pp. 7274--7283.

\bibitem{ZhangWLWR17}
P.~Zhang, D.~Wang, H.~Lu, H.~Wang, and X.~Ruan, ``Amulet: Aggregating
  multi-level convolutional features for salient object detection,'' in
  \emph{ICCV}, 2017, pp. 202--211.

\bibitem{wang2019progressive}
B.~Wang, Q.~Chen, M.~Zhou, Z.~Zhang, X.~Jin, and K.~Gai, ``Progressive feature
  polishing network for salient object detection,'' in \emph{AAAI}, 2020, pp.
  12\,128--12\,135.

\bibitem{wei2019f3net}
J.~Wei, S.~Wang, and Q.~Huang, ``F3net: Fusion, feedback and focus for salient
  object detection,'' in \emph{AAAI}, 2020, pp. 123\,221--12\,328.

\bibitem{PanYLM17caption}
Y.~Pan, T.~Yao, H.~Li, and T.~Mei, ``Video captioning with transferred semantic
  attributes,'' in \emph{CVPR}, 2017, pp. 984--992.

\bibitem{ZhangFU16drive}
Z.~Zhang, S.~Fidler, and R.~Urtasun, ``Instance-level segmentation for
  autonomous driving with deep densely connected mrfs,'' in \emph{CVPR}, 2016,
  pp. 669--677.

\bibitem{XuPCYH16interactive}
N.~Xu, B.~L. Price, S.~Cohen, J.~Yang, and T.~S. Huang, ``Deep interactive
  object selection,'' in \emph{CVPR}, 2016, pp. 373--381.

\bibitem{XieT17edge}
S.~Xie and Z.~Tu, ``Holistically-nested edge detection,'' \emph{IJCV}, vol.
  125, no. 1-3, pp. 3--18, 2017.

\bibitem{ZhugeYZL18edge}
Y.~Zhuge, G.~Yang, P.~Zhang, and H.~Lu, ``Boundary-guided feature aggregation
  network for salient object detection,'' \emph{IEEE SPL}, vol.~25, no.~12, pp.
  1800--1804, 2018.

\bibitem{zhao2019egnet}
J.-X. Zhao, J.-J. Liu, D.-P. Fan, Y.~Cao, J.~Yang, and M.-M. Cheng, ``{EGNet:
  Edge guidance network for salient object detection},'' in \emph{ICCV}, 2019,
  pp. 8779--8788.

\bibitem{wu2019stacked}
Z.~Wu, L.~Su, and Q.~Huang, ``Stacked cross refinement network for edge-aware
  salient object detection,'' in \emph{ICCV}, 2019, pp. 7264--7273.

\bibitem{desingh2013BMCV}
K.~Desingh, K.~Krishna, D.~Rajanand, and C.~Jawahar, ``Depth really matters:
  Improving visual salient region detection with depth,'' in \emph{BMVC}, 2013,
  pp. 1--11.

\bibitem{ciptadi2013BMCV}
A.~Ciptadi, T.~Hermans, and J.~M.~Rehg, ``An in depth view of saliency,'' in
  \emph{BMVC}, 2013, pp. 1--11.

\bibitem{qu2017DF}
L.~Qu, S.~He, J.~Zhang, J.~Tian, Y.~Tang, and Q.~Yang, ``{RGBD} salient object
  detection via deep fusion,'' \emph{IEEE TIP}, vol.~26, no.~5, pp. 2274--2285,
  2017.

\bibitem{cong2019DGTM}
R.~{Cong}, J.~{Lei}, H.~{Fu}, J.~{Hou}, Q.~{Huang}, and S.~{Kwong}, ``Going
  from {RGB} to {RGBD} saliency: A depth-guided transformation model,''
  \emph{IEEE TCYBERNETICS}, pp. 1--13, 2019.

\bibitem{Shigematsu2017BED}
R.~Shigematsu, D.~Feng, S.~You, and N.~Barnes, ``Learning {RGB-D} salient
  object detection using background enclosure, depth contrast, and top-down
  features,'' in \emph{CVPRW}, 2017, pp. 2749--2757.

\bibitem{He2016resnet}
K.~He, X.~Zhang, S.~Ren, and J.~Sun, ``Deep residual learning for image
  recognition,'' in \emph{CVPR}, 2016, pp. 770--778.

\bibitem{han2018CTMF}
J.~Han, H.~Chen, N.~Liu, C.~Yan, and X.~Li, ``{CNNs-Based} {RGB-D} saliency
  detection via cross-view transfer and multiview fusion,'' \emph{IEEE
  TCYBERNETICS}, vol.~48, no.~11, pp. 3171--3183, 2018.

\bibitem{Zhang2020UCNet}
J.~Zhang, D.-P. Fan, Y.~Dai, S.~Anwar, F.~Sadat~Saleh, T.~Zhang, and N.~Barnes,
  ``{UC-Net: Uncertainty Inspired RGB-D Saliency Detection via Conditional
  Variational Autoencoders},'' in \emph{CVPR}, 2020, pp. 8582--8591.

\bibitem{wang2020synergistic}
Y.~Wang, Y.~Li, J.~H. Elder, H.~Lu, and R.~Wu, ``{Synergistic saliency and
  depth prediction for RGB-D saliency detection},'' in \emph{ACCV}, 2021.

\bibitem{Fu2020JLDCF}
K.~F. Fu, D.-P. Fan, G.-P. Ji, and Q.~Zhao, ``{JL-DCF: Joint Learning and
  Densely-Cooperative Fusion Framework for RGB-D Salient Object Detection},''
  in \emph{CVPR}, 2020, pp. 3052--3062.

\bibitem{Luo_2020_ECCV}
A.~Luo, X.~Li, F.~Yang, Z.~Jiao, H.~Cheng, and S.~Lyu, ``{Cascade Graph Neural
  Networks for RGB-D Salient Object Detection},'' in \emph{ECCV}, 2020, pp.
  346--364.

\bibitem{zhao2020single}
X.~Zhao, L.~Zhang, Y.~Pang, H.~Lu, and L.~Zhang, ``{A Single Stream Network for
  Robust and Real-time RGB-D Salient Object Detection},'' in \emph{ECCV}, 2020,
  pp. 646--662.

\bibitem{li2020rgbd}
C.~Li, R.~Cong, Y.~Piao, Q.~Xu, and C.~C. Loy, ``{RGB-D Salient Object
  Detection with Cross-Modality Modulation and Selection},'' in \emph{ECCV},
  2020, pp. 225--241.

\bibitem{zhou2020rgbd}
T.~Zhou, D.-P. Fan, M.-M. Cheng, J.~Shen, and L.~Shao, ``{RGB-D Salient Object
  Detection: A Survey},'' \emph{IEEE TMM}, 2021.

\bibitem{liu2018rfb}
S.~Liu, D.~Huang, and Y.~Wang, ``Receptive field block net for accurate and
  fast object detection,'' in \emph{ECCV}, 2018, pp. 404--419.

\bibitem{2019huAcnet}
X.~{Hu}, K.~{Yang}, L.~{Fei}, and K.~{Wang}, ``{ACNet: Attention Based Network
  to Exploit Complementary Features for RGBD Semantic Segmentation},'' in
  \emph{ICIP}, 2019, pp. 1440--1444.

\bibitem{chen2017photographic}
Q.~Chen and V.~Koltun, ``Photographic image synthesis with cascaded refinement
  networks,'' in \emph{CVPR}, 2017, pp. 1511--1520.

\bibitem{WangBZZL17}
T.~Wang, A.~Borji, L.~Zhang, P.~Zhang, and H.~Lu, ``A stagewise refinement
  model for detecting salient objects in images,'' in \emph{ICCV}, 2017, pp.
  4039--4048.

\bibitem{deng2018r3net}
Z.~Deng, X.~Hu, L.~Zhu, X.~Xu, J.~Qin, G.~Han, and P.-A. Heng, ``{R3Net:
  Recurrent residual refinement network for saliency detection},'' in
  \emph{IJCAI}, 2018, pp. 684--690.

\bibitem{Woo2018CBAM}
S.~Woo, J.~Park, J.-Y. Lee, and I.~So~Kweon, ``{CBAM}: Convolutional block
  attention module,'' in \emph{ECCV}, 2018, pp. 3--19.

\bibitem{oquab2015GMP}
M.~{Oquab}, L.~{Bottou}, I.~{Laptev}, and J.~{Sivic}, ``Is object localization
  for free? - weakly-supervised learning with convolutional neural networks,''
  in \emph{CVPR}, 2015, pp. 685--694.

\bibitem{steiner2019pytorch}
B.~Steiner, Z.~DeVito, S.~Chintala, S.~Gross, A.~Paszke, F.~Massa, A.~Lerer,
  G.~Chanan, Z.~Lin, E.~Yang \emph{et~al.}, ``{PyTorch}: An imperative style,
  high-performance deep learning library,'' in \emph{NIPS}, 2019, pp.
  8024--8035.

\bibitem{KrizhevskySH2012Imagenet}
A.~Krizhevsky, I.~Sutskever, and G.~E. Hinton, ``Imagenet classification with
  deep convolutional neural networks,'' in \emph{NIPS}, 2012, pp. 1106--1114.

\bibitem{KingmaB2014adam}
D.~P. Kingma and J.~Ba, ``Adam: {A} method for stochastic optimization,'' in
  \emph{ICLR}, 2015.

\bibitem{li2014LFSD}
N.~Li, J.~Ye, Y.~Ji, H.~Ling, and J.~Yu, ``Saliency detection on light field,''
  in \emph{CVPR}, 2014, pp. 2806--2813.

\bibitem{zhu2017SSD}
C.~Zhu and G.~Li, ``A three-pathway psychobiological framework of salient
  object detection using stereoscopic technology,'' in \emph{CVPRW}, 2017, pp.
  3008--3014.

\bibitem{ZhuLGWW17MB}
C.~Zhu, G.~Li, X.~Guo, W.~Wang, and R.~Wang, ``A multilayer backpropagation
  saliency detection algorithm based on depth mining,'' in \emph{CAIP}, 2017,
  pp. 14--23.

\bibitem{cong2016DCMC}
R.~Cong, J.~Lei, C.~Zhang, Q.~Huang, X.~Cao, and C.~Hou, ``Saliency detection
  for stereoscopic images based on depth confidence analysis and multiple cues
  fusion,'' \emph{IEEE SPL}, vol.~23, no.~6, pp. 819--823, 2016.

\bibitem{piao2020a2dele}
Y.~Piao, Z.~Rong, M.~Zhang, W.~Ren, and H.~Lu, ``{A2dele: Adaptive and
  Attentive Depth Distiller for Efficient RGB-D Salient Object Detection},'' in
  \emph{CVPR}, 2020, pp. 9060--9069.

\bibitem{zhang2020select}
M.~Zhang, W.~Ren, Y.~Piao, Z.~Rong, and H.~Lu, ``{Select, Supplement and Focus
  for RGB-D Saliency Detection},'' in \emph{CVPR}, 2020, pp. 3472--3481.

\bibitem{liang2018CDB}
F.~Liang, L.~Duan, W.~Ma, Y.~Qiao, Z.~Cai, and L.~Qing, ``Stereoscopic saliency
  model using contrast and depth-guided-background prior,''
  \emph{Neurocomputing}, vol. 275, pp. 2227--2238, 2018.

\bibitem{ren2015GP}
J.~Ren, X.~Gong, L.~Yu, W.~Zhou, and M.~Ying~Yang, ``Exploiting global priors
  for {RGB-D} saliency detection,'' in \emph{CVPRW}, 2015, pp. 25--32.

\bibitem{song2017MDSF}
H.~Song, Z.~Liu, H.~Du, G.~Sun, O.~Le~Meur, and T.~Ren, ``Depth-aware salient
  object detection and segmentation via multiscale discriminative saliency
  fusion and bootstrap learning,'' \emph{IEEE TIP}, vol.~26, no.~9, pp.
  4204--4216, 2017.

\bibitem{Fan2017Smeasure}
D.-P. Fan, M.-M. Cheng, Y.~Liu, T.~Li, and A.~Borji, ``Structure-measure: A new
  way to evaluate foreground maps,'' in \emph{ICCV}, 2017, pp. 4548--4557.

\bibitem{Fan2018Emeasure}
D.-P. Fan, C.~Gong, Y.~Cao, B.~Ren, M.-M. Cheng, and A.~Borji,
  ``Enhanced-alignment measure for binary foreground map evaluation,'' in
  \emph{IJCAI}, 2018, pp. 698--–704.

\bibitem{Ach2009Fmeasure}
R.~Achanta, S.~Hemami, F.~Estrada, and S.~Susstrunk, ``Frequency-tuned salient
  region detection,'' in \emph{CVPR}, 2009, pp. 1597--1604.

\bibitem{simonyan2014vgg}
K.~Simonyan and A.~Zisserman, ``Very deep convolutional networks for
  large-scale image recognition,'' \emph{arXiv preprint arXiv:1409.1556}, 2014.

\bibitem{Liu_2018_PiCAN}
N.~Liu, J.~Han, and M.-H. Yang, ``{PiCANet: Learning Pixel-Wise Contextual
  Attention for Saliency Detection},'' in \emph{CVPR}, 2018, pp. 3089--3098.

\bibitem{yang2016top}
J.~Yang and M.-H. Yang, ``Top-down visual saliency via joint crf and dictionary
  learning,'' \emph{IEEE TPAMI}, vol.~39, no.~3, pp. 576--588, 2016.

\bibitem{wang2019pyramid}
W.~Wang, S.~Zhao, J.~Shen, S.~C. Hoi, and A.~Borji, ``Salient object detection
  with pyramid attention and salient edges,'' in \emph{CVPR}, 2019, pp.
  1448--1457.

\bibitem{zeng2019joint}
Y.~Zeng, Y.~Zhuge, H.~Lu, and L.~Zhang, ``Joint learning of saliency detection
  and weakly supervised semantic segmentation,'' in \emph{ICCV}, 2019, pp.
  7223--7233.

\bibitem{otsu1979threshold}
N.~Otsu, ``A threshold selection method from gray-level histograms,''
  \emph{IEEE SMC}, vol.~9, no.~1, pp. 62--66, 1979.

\bibitem{KrahenbuhlK11crf}
P.~Kr{\"{a}}henb{\"{u}}hl and V.~Koltun, ``Efficient inference in fully
  connected crfs with gaussian edge potentials,'' in \emph{NIPS}, 2011, pp.
  109--117.

\end{thebibliography}

%
%




\end{document}